\documentclass[10pt,twocolumn,letterpaper]{article}

\usepackage[pagenumbers]{cvpr} % To force page numbers, e.g. for an arXiv version

\usepackage[dvipsnames]{xcolor}

\definecolor{cvprblue}{rgb}{0.21,0.49,0.74}
\usepackage[pagebackref,breaklinks,colorlinks,citecolor=cvprblue]{hyperref}
\usepackage{lipsum}
\usepackage{multirow}
\usepackage{pifont}
\usepackage{paralist}
\usepackage{comment}
\usepackage[hang,flushmargin]{footmisc}
\usepackage[dvipsnames]{xcolor}

\newcommand{\mycaption}[2]{\caption{\textbf{#1.}~#2}}
\newcommand{\cmark}{\ding{51}}
\newcommand{\xmark}{\ding{55}}
\usepackage[toc,page,header]{appendix}
\usepackage{minitoc}
\def\paperID{2889} % *** Enter the Paper ID here
\def\confName{CVPR}
\def\confYear{2024}

\newcommand\blfootnote[1]{%
  \begingroup
  \renewcommand\thefootnote{}\footnote{#1}%
  \addtocounter{footnote}{-1}%
  \endgroup
}

\begin{document}

\title{Panda-70M: Captioning 70M Videos with Multiple Cross-Modality Teachers}

%%%%%%%%% AUTHORS - PLEASE UPDATE
\author{
Tsai-Shien Chen$^{1,2,*}$ \quad Aliaksandr Siarohin$^1$ \quad Willi Menapace$^{1,3,*}$ \quad Ekaterina Deyneka$^1$ \\
Hsiang-wei Chao$^1$ \quad Byung Eun Jeon$^1$ \quad Yuwei Fang$^1$ \quad Hsin-Ying Lee$^1$ \quad Jian Ren$^1$ \\
Ming-Hsuan Yang$^{2}$ \quad Sergey Tulyakov$^1$ \\
{\normalsize $^1$Snap Inc. \quad $^2$University of California, Merced \quad $^3$University of Trento} \\
{\small \url{https://snap-research.github.io/Panda-70M}}
}

\doparttoc % Tell to minitoc to generate a toc for the parts
\faketableofcontents % Run a fake tableofcontents command for the partocs

\twocolumn[{
\renewcommand\twocolumn[1][]{#1}
\maketitle
\begin{center}
    \centering
    \captionsetup{type=figure}
    \vspace{-8mm}
    \includegraphics[width=\textwidth]{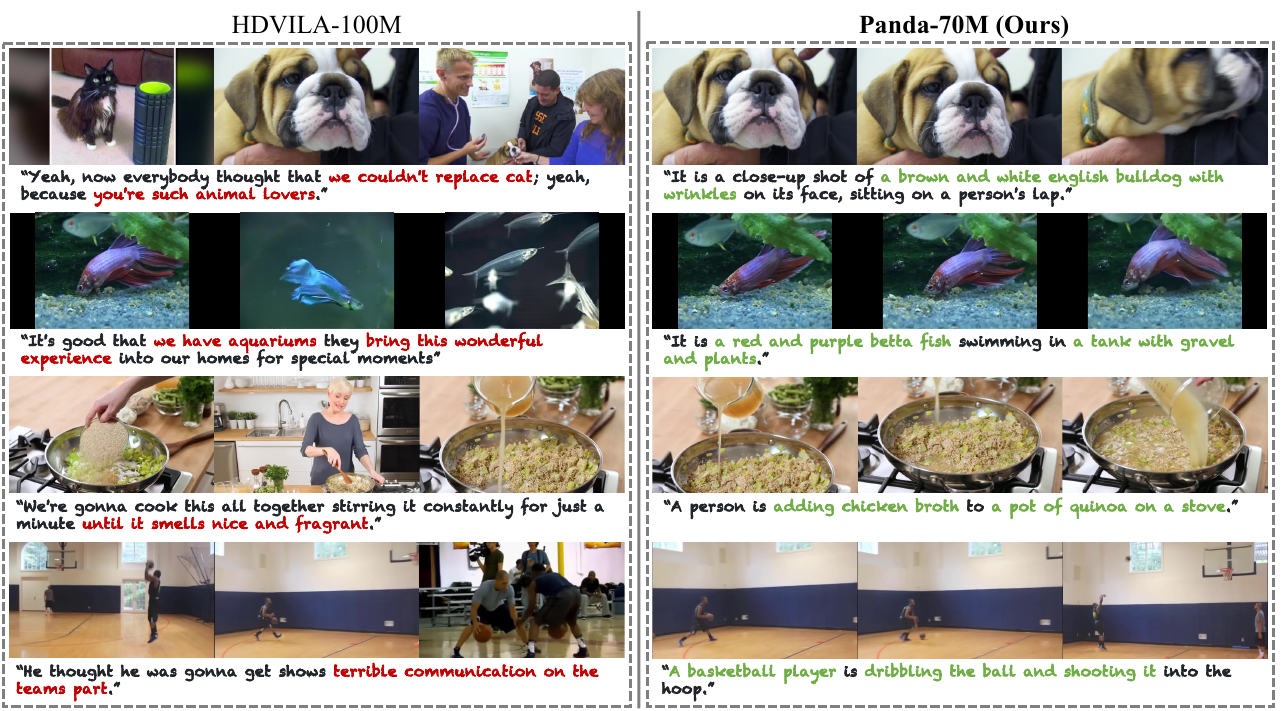}
    \vspace{-7mm}
    \mycaption{Comparison of Panda-70M to the existing large-scale video-language datasets}{
        We introduce Panda-70M, a large-scale video dataset with captions that are annotated by multiple cross-modality vision-language models.
        %
        %Leveraging from multi-modality information,
        Compared to text annotations in existing dataset~\cite{hdvila}, captions in Panda-70M more precisely describe the main object and action in videos (highlighted in {\color{YellowGreen}green}).
        Besides, videos in Panda-70M are semantically coherent, high-resolution, and free from watermarks.
        More samples can be found in Appendix~\ref{app:visualization}.
        %
        %(a) Snap-70M equips several superior advantages over the existing datasets, including grammatically correct and informative caption label, semantics consistency, high resolution and lack of watermark.
        %(b) With high quality video-caption pair, Snap-70M can facilitate V2T/T2V retrieval, video captioning, and T2V generation downstream tasks.
    }
    \vspace{+1mm}
    \label{fig:teaser}
\end{center}
}]

\blfootnote{\hspace{-0.21cm} $^*$ This work was done while interning at Snap.}
\begin{abstract}
\vspace{-8mm}
% We introduce Panda-70M, a large-scale video-captioning dataset annotated by multi-modalities vision-language models.
% %
% The video clips in Panda-70M are extracted from 3.8M long HD-VILA-100M dataset and feature several advantages: semantics coherence, high-resolution ($>$720p), and watermark-free.
% %
% To annotate the videos with high-quality captions, we propose an automatic pipeline which can benefit from multimodal information input, like video description, subtitles, and individual video frames.
% %
% Specifically, we employ multiple-modalities teacher models, composed by image captioning models and image/video visual question answering models with additional text inputs, to generate several caption candidates for a video clip.
% %
% Subsequently, we fine-tune a fine-grained retrieval model to select the most precise caption as the annotation.
% %
% To purse infinite-scale video-captioning dataset, we design a multimodal student captioning framework and train this model to distill the knowledge from multiple teachers.
% %and show that the student can outperform any teacher.
% %
% We evaluate the effectiveness of the proposed dataset on three downstream tasks: video captioning, video and text retrieval, and text-driven video generation.
% %

The quality of the data and annotation upper-bounds the quality of a downstream model.
While there exist large text corpora and image-text pairs, high-quality video-text data is much harder to collect. 
First of all, manual labeling is more time-consuming, as it requires an annotator to watch an entire video.
Second, videos have a temporal dimension, consisting of several scenes stacked together, and showing multiple actions.
Accordingly, to establish a video dataset with high-quality captions, we propose an automatic approach leveraging multimodal inputs, such as textual video description, subtitles, and individual video frames.
Specifically, we curate 3.8M high-resolution videos from the publicly available HD-VILA-100M dataset. 
We then split them into semantically consistent video clips, and apply multiple cross-modality teacher models to obtain captions for each video.
% We then split them into semantically consistent scenes and applied multiple teacher models including image captioning, and image- or video-question-answering. Each teacher provided a caption for a video. 
%
Next, we finetune a retrieval model on a small subset where the best caption of each video is manually selected and then employ the model in the whole dataset to select the best caption as the annotation.
In this way, we get 70M videos paired with high-quality text captions. We dub the dataset as Panda-70M.
%
% Using knowledge distillation from multiple teachers we obtained a student model that labelled the remainder of the data.
%
We show the value of the proposed dataset on three downstream tasks: video captioning, video and text retrieval, and text-driven video generation. The models trained on the proposed data score substantially better on the majority of metrics across all the tasks.

\end{abstract}
\begin{table*}[t]
    \centering
    \small
    \mycaption{Comparison of Panda-70M and other video-language datasets}{
        We split the datasets into two groups: the group at the top is annotated by ASR, and the group at the bottom is labeled with captions. %\sergey{manually?} \tschen{yes}
        %
        % To scale up the video dataset with caption annotations, we propose an automatic pipeline that makes use of multimodal information to generate high-quality captions.
        %
        % Our Panda-70M has video samples with higher resolution, filtered out watermarks. \tschen{comment out this line as we remove watermark column in the table}
        %
    }
    \label{tab:dataset}
    \setlength\tabcolsep{4.8pt}
  \vspace{-3mm}
  \begin{tabular}{lccccccccc}
    \toprule
    Dataset & Year & Text & Domain & \#Videos & \multicolumn{2}{c}{Avg/Total video len} & Avg text len & Resolution \\
    \midrule
    HowTo100M~\cite{howto100m}     & 2019 & ASR               & Open    & 136M   & 3.6s  & 134.5Khr & 4.0 words  & 240p  \\
    ACAV~\cite{acav}               & 2021 & ASR               & Open    & 100M   & 10.0s & 277.7Khr & -          & -     \\
    YT-Temporal-180M~\cite{merlot} & 2021 & ASR               & Open    & 180M   & -     & -        & -          & -     \\
    HD-VILA-100M~\cite{hdvila}     & 2022 & ASR               & Open    & 103M   & 13.4s & 371.5Khr & 32.5 words & 720p  \\
    \midrule
    MSVD~\cite{msvd}               & 2011 & Manual caption    & Open    & 1970   & 9.7s  & 5.3h     & 8.7 words  & -     \\
    LSMDC~\cite{lsmdc}             & 2015 & Manual caption    & Movie   & 118K   & 4.8s  & 158h     & 7.0 words  & 1080p \\
    MSR-VTT~\cite{msrvtt}          & 2016 & Manual caption    & Open    & 10K    & 15.0s & 40h      & 9.3 words  & 240p  \\
    DiDeMo~\cite{didemo}           & 2017 & Manual caption    & Flickr  & 27K    & 6.9s  & 87h      & 8.0 words  & -     \\
    ActivityNet~\cite{activitynet} & 2017 & Manual caption    & Action  & 100K   & 36.0s & 849h     & 13.5 words & -     \\
    YouCook2~\cite{youcook2}       & 2018 & Manual caption    & Cooking & 14K    & 19.6s & 176h     & 8.8 words  & -     \\
    VATEX~\cite{vatex}             & 2019 & Manual caption    & Open    & 41K    & $\sim$10s & $\sim$115h & 15.2 words & - \\
    \midrule
    \textbf{Panda-70M (Ours)}      & 2024 & Automatic caption & Open    & 70.8M  & 8.5s  & 166.8Khr & 13.2 words & 720p \\
    \bottomrule
    \end{tabular}
    \vspace{-3.5mm}
\end{table*}

\section{Introduction}
\label{sec:introduction}
\vspace{-2mm}

We enter an era where the size of computing and data are indispensable for large-scale multimodal learning.
Most breakthroughs are achieved by large-scale computing infrastructure, large-scale models, and large-scale data.
Due to these integral components, we have powerful text-to-image~\cite{dalle,imagen,ldm,ediffi,parti} and image-to-text models~\cite{blip,chatgpt4,flamingo,llava}.
Scaling the model size or the compute is challenging and expensive; however, it requires a finite amount of engineering time.
Scaling the data is relatively more challenging, as it takes time for a human to analyze each sample.

Especially, compared to image-text pairs~\cite{cc12m,coyo,laion}, video-text pairs are even harder to obtain.
First, annotating videos is more time-consuming, as an annotator needs to watch the entire video before labeling.
Second, videos often contain multiple scenes stitched together and consist of temporally varying content.
Finally, meta-information, such as subtitles, video description, and voice-over, is often too broad or not correctly aligned in time or cannot precisely describe a video.
For example, several 100M-scale datasets, such as HD-VILA-100M~\cite{hdvila} and HowTo100M~\cite{howto100m}, are annotated by automatic speech recognition (ASR).
However, as shown in Figure~\ref{fig:teaser}, the subtitles usually fail to include the main content and action presented in the video.
This limits the value of such datasets for multimodal training.
%, {\color{red} upper-bounding their scores as we show in the Sec. X. Figure \ref{}, left shows several examples of such data.}
We summarize the datasets available to the community in Table~\ref{tab:dataset}.
Some are low-resolution, some are annotated by ASR, some contain data from a limited domain, some are small-scale, and some offer short captions. 

In this work, we present a large-scale dataset containing 70M video clips with caption annotations.
It includes high-resolution videos from an open domain with rich captions averaging 13.2 words per caption.
%
% Our approach to collecting such a dataset consists of both manual and automatic annotation.
While manually annotating 70M videos is prohibitively expensive, we opt for automatic annotation.
Our key insight is that a video typically comes with information from several modalities that can assist automatic captioning.
This includes the title, description, subtitles of the video, individual static frames, and the video itself. 
The value of this data cannot be fully maximized when only partially used.
In comparison, we propose to utilize different combinations of multimodal data as inputs to various cross-modality captioning models.
% observe a significant improvement when different combinations of modalities are employed by an ensemble of different captioning models.
%
% In particular, we propose to utilize different combinations of the multimodal data as inputs to a number of cross-modality teachers.
%
% We argue that different combinations of such data are useful to describe different videos. 
%
To substantiate this idea, we conduct a numerical analysis based on a human evaluation (the details are provided in Appendix~\ref{app:user_study_all_good}).
If we use multiple cross-modality models to caption some video samples and evaluate the results by showing them to humans, we see that there is no single model able to generate good captions for more than $31\%$ of videos. However, if we jointly collect all the captions from different models, we observe that $84.7\%$ of videos can be annotated with at least one good caption.
% \tschen{Is it clearer? Please check! Thanks}
% \sergey{35\% the right number? in Fig 3 the highest is VideoLLAMA it's 23\%}
% \sergey{this number is not in the figure, should we add?}
% \tschen{they are based on different user study}
% \tschen{this one is from "select all good" user study => we can respectively analyze how each teacher model works from this study}
% \tschen{In comparison, Figure 3 is from "select the best" user study}

\begin{figure*}[t]
    \centering
    \includegraphics[width=\textwidth]{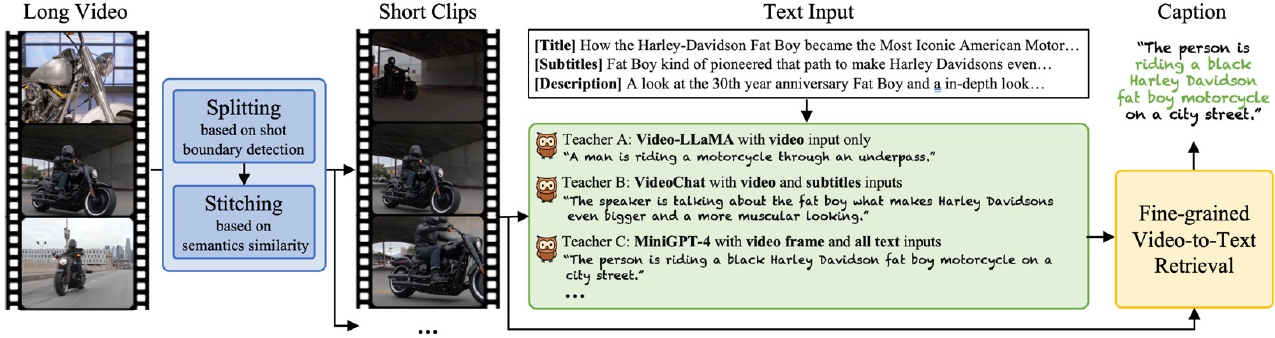}
    \vspace{-7mm}
    \mycaption{Video captioning pipeline}{
        Given a long video, we first split it into several semantically coherent clips.
        Subsequently, we utilize a number of teacher models with different multimodal inputs to generate multiple captions for a video clip.
        Lastly, we finetune a fine-grained retrieval model to select the caption that best describes the video clip as the annotation.
    }
    \vspace{-4mm}
    \label{fig:pipeline}
\end{figure*}
%%%%%%%%%%%%%%%%%%%%%%%%%%%%%%%

To establish the dataset with this mindset, we begin by using 3.8M high-resolution long videos collected from HD-VILA-100M~\cite{hdvila} and process them through the following three steps.
First, we design a semantics-aware video splitting algorithm to cut long videos into semantically consistent clips while striking the balance between semantics coherence and the duration of the video clips.
% In total, we get 70M semantically consistent video clips. 
%
Second, we use a range of cross-modality teacher models, including image captioning models~\cite{blipv2} and image/video visual-question-answering (VQA) models~\cite{minigpt4,videollama,videochat} with additional text inputs, such as video description and subtitles, to predict several candidate captions for a clip.
Lastly, we collect a 100K video subset, where human annotators act as an oracle to select the best caption for each video. We use this dataset to finetune a fine-grained video-to-text retrieval model~\cite{unmaskedteacher} which is then applied to the whole dataset to select the most precise caption as the annotation.
Running multiple teacher models is computationally expensive and time-consuming. To pursue efficient video captioning at scale in the future, we train a student model to distill the knowledge from the teachers.
The student model adopts a two-branch architecture which can take both visual and textual inputs to benefit the captioning from multimodal information.

Extensive experiments demonstrate that pretraining with the proposed Panda-70M\footnote{We call our dataset Panda, drawing an analogy to Panda Po, who learns from multiple martial arts teachers.} can benefit several downstream tasks, including video captioning, video and text retrieval, and text-to-video generation.
We also show that training a student model in a knowledge distillation manner facilitates learning a strong student model which can outperform any teacher model by more than $7.7\%$ preference ratio as in Table~\ref{tab:captioning}, where the performance can be further enhanced by additional text inputs, like video description and subtitles.
%\sergey{by what margin?}\tschen{wait for the user study}

%\sergey{not 100\% sure we need to list the contributions again}
%\tschen{agreed, information here is redundant}
\begin{comment}
The main contributions of this paper are:
\begin{compactitem}
\item We introduce Panda-70M, a video-captioning dataset that includes semantically consistent, high resolution, and watermark-free video samples.
\item We propose an automatic annotation pipeline that fully utilizes multimodal information, such as video description and subtitles, for captioning.
% \item We propose a dual-branch video captioning model which can take both visual and textual inputs to advance the captioning capability.
\item Extensive experiments show that pretraining with Panda-70M can yield substantial improvement on three downstream tasks.
\end{compactitem}
\end{comment}

\vspace{-1.9mm}
\section{Related Work}
\label{sec:related_work}
\vspace{-1.9mm}
%We introduce Panda-70M, a large-scale vision-language dataset which is automatically annotated by multiple vision-language teacher models.

\noindent \textbf{Vision-Language Datasets.}
Training with millions or even billions of image-text pairs~\cite{clip,align,florence,coyo,laion} has been shown to be effective in learning powerful image foundation models~\cite{transformers,mae,beit,flamingo,coca,imagebind}.
With this work, our goal is to build a large video-language dataset containing rich captions. We compare related datasets in Table~\ref{tab:dataset}.
Several precedent video-language datasets~\cite{msvd,lsmdc,msrvtt,didemo,activitynet,youcook2,vatex} contain data tackling various tasks, such as action recognition, video understanding, VQA, and retrieval.
However, manually annotating data is costly and limits the scale of such datasets (typically they contain less than 120K samples).
To alleviate the lack of data, the works of~\cite{howto100m,merlot,hdvila} propose to automatically annotate data with subtitles, generated by ASR.
While this approach significantly increases the dataset scale reaching 100M of samples, the subtitles, unfortunately, do not precisely describe the main video content, as shown in Figure~\ref{fig:teaser}.
%
% Additionally, the videos are split based on the alignment of subtitle sentences, which would lead to semantically inconsistent clips and makes the text labels hard to accurately summarize the video.
%
% An alternative approach is to collect stock videos with user specified description from the web.
%
% Captions is such datasets are sometimes imprecise, generic or irrelevant to the videos and these videos usually contain watermarks.
%
% In comparison, Panda-70M is consists of semantically consistent, high-resolution, and watermark-free video samples.
%
% Moreover, our captions are obtained by using multiple teachers with multimodal inputs, which can usually capture the object and the motion presented in the video. 
% leveraging from multiple cross-modality teacher models, the caption labels can usually capture the critical object and motion presented in the videos. 
In comparison, in this work, we propose an automatic captioning pipeline with the inputs of multimodal data that enables us to scale up the dataset of high-quality video-caption pairs to a 70M scale.
%
%The algorithm make scaling up the video dataset with high-quality captions practical
%while making good use of multimodal information to generate high-quality caption annotations.

\noindent \textbf{Vision-Language Models}
learn the correlation between visual data (images or videos) and linguistic signals (words or sentences) and can be applied to several downstream applications, including text-driven image or video generation~\cite{dalle,imagen,ldm,ediffi,parti,magicvideo,imagenvideo,makeavideo,videoldm}, captioning~\cite{blip,blipv2,flamingo,llava,vid2seq,textkg}, VQA~\cite{chatgpt4,minigpt4,minigptv2,videollama,videochat} and retrieval~\cite{clip4clip,vindlu,unmaskedteacher}.
We utilize several vision-language models for the annotation of Panda-70M.
BLIP-2~\cite{blipv2} introduces an efficient vision-language pretraining that can facilitate image captioning.
We use BLIP-2 as one of the teachers and input a randomly sampled video frame for captioning.
MiniGPT-4~\cite{minigpt4} is an image VQA model that learns a projection layer to align a large language model (LLM) and a visual encoder.
In addition to a video frame, we also input a prompt with extra text information, such as video description and subtitles, and ask the model to summarize all multimodal inputs.
For the video modality, Video-LLaMA~\cite{videollama} and VideoChat~\cite{videochat} are both video VQA models and learn to extract LLM-compatible visual embeddings.
We use both models and ask them to caption a video with prompt input.
Besides, Unmasked Teacher~\cite{unmaskedteacher} is a video foundation model which can facilitate video understanding.
We finetune it to implement fine-grained retrieval and use it to select the more precise caption as the annotation.
%Finally, we also explore video-captioning model that take multimodal input.

\noindent \textbf{Video Annotation through Multi-modal Models.}
With the aforementioned development on vision-language models, some concurrent works~\cite{videofactory,internvid,svd} also leverage these models for video captioning.
VideoFactory~\cite{videofactory} employs BLIP-2~\cite{blipv2} to caption video clips.
However, as reported in Appendix~\ref{app:user_study_all_good}, the performance of a single BLIP-2 model is suboptimal.
More similar to our captioning pipeline, InternVid~\cite{internvid} and Stable Video Diffusion~\cite{svd} also use multiple captioning models which are followed by an LLM for summarization.
In practice, we found the LLM would propagate errors from noisy outputs of vision-language models.
\vspace{-6mm}
\section{Methodology}
\label{sec:methodology}
\vspace{-1.5mm}

To build Panda-70M, we utilize 3.8M high-resolution long videos collected from HD-VILA-100M~\cite{hdvila}.
We then split them into 70.8M semantically coherent clips as described in Section~\ref{sec:splitting}.
Section~\ref{sec:teacher} shows how multiple cross-modality teacher models are used to generate a set of candidate caption annotations.
Next, we finetune a fine-grained retrieval model to select the most accurate caption as detailed in Section~\ref{sec:v2t_retrieval}.
Finally, in Section~\ref{sec:student}, we describe our approach to training a student captioning model using Panda-70M.
The high-level view of our approach is shown in Figure~\ref{fig:pipeline}.

\vspace{-1mm}
\subsection{Semantics-aware Video Splitting}
\label{sec:splitting}
\vspace{-1.5mm}

A desired video sample in a video-captioning dataset should have two somewhat contradictory characteristics.
On the one hand, the video should be semantically consistent, so the video samples can better benefit the downstream tasks, such as action recognition, and the caption can also more accurately express its semantics content without ambiguity.
On the other hand, the video cannot be too short or fragmentary to contain meaningful motion content, which is beneficial to tasks, like video generation.

To achieve both goals, we design a two-stage semantics-aware splitting algorithm to cut a long video into semantically coherent clips.
In the first stage, we split the video based on shot boundary detection~\cite{pyscenedetect}, as the semantics often change when a new scene starts.
In the second stage, we stitch adjacent clips if they are incorrectly separated by the first stage, ensuring the videos do not end up being too short.
To do so, we use ImageBind~\cite{imagebind} to extract embeddings of video frames and merge the adjacent clips if the frame embeddings from two clips are similar.
We also implement additional procedures to handle: 1) long videos without any cut-scenes, 2) videos using complex transitions, such as fade-in and fade-out effects, which are not usually detected as cut-scenes, and 3) removal of redundant clips to increase the diversity of the dataset.
More details of the splitting algorithm are in Appendix~\ref{app:splitting}.
Notably, while our dataset focuses on fine-grained video-text pairs with consistent semantics, users can still acquire long videos with multiple cut-scenes by concatenating consecutive clips and captions, as these clips are split from the same long video.

To quantitatively verify the semantic consistency of a video clip, we introduce Max Running LPIPS, which highlights the most significant perceptual change within a video clip.
Formally, given an $n$-second video clip, we subsample the video frames each second and denote the keyframes as $\{f_1, ..., f_n\}$. The Max Running LPIPS is formulated as:
\begin{equation}
    \label{eq:max_running_LPIPS}
    \vspace{-1mm}
    \max(\{\text{LPIPS}(f_i, f_{i+1}) \mid i \in [1,n-1]\}).
\end{equation}
where LPIPS$(\cdot,\cdot)$ is the perceptual similarity~\cite{lpips} of two images.
As in Table~\ref{tab:splitting}, our splitting achieves a better semantics consistency than the splitting based on the alignment of subtitles sentences~\cite{hdvila,howto100m}, while maintaining longer video length than the vanilla shot boundary detection~\cite{pyscenedetect}.

\begin{table}[t]
    \centering
    \small
    \mycaption{Comparison of splitting algorithms}{
        We split 1K long videos by three algorithms and test the semantics consistency of the output clips by the proposed Max Running LPIPS.
        Our splitting strikes a better balance for the trade-off between semantics consistency and clip length.
    }
    \label{tab:splitting}

    \setlength\tabcolsep{4pt}
    \vspace{-3mm}
    \begin{tabular}{lcc}
    \toprule
    Method & Max running LPIPS$\downarrow$ & Avg Video Len \\
    \midrule
    Sub. Align~\cite{howto100m,hdvila}  & 0.408 & 11.8s \\
    PySceneDetect~\cite{pyscenedetect}  & 0.247 & 4.1s \\
    \midrule
    Our Splitting                       & 0.256 & 7.9s \\
    \bottomrule
    \end{tabular}
    %\vspace{-1mm}
\end{table}
\begin{figure}[t]
    \centering
    \vspace{-2mm}
    \includegraphics[width=\linewidth]{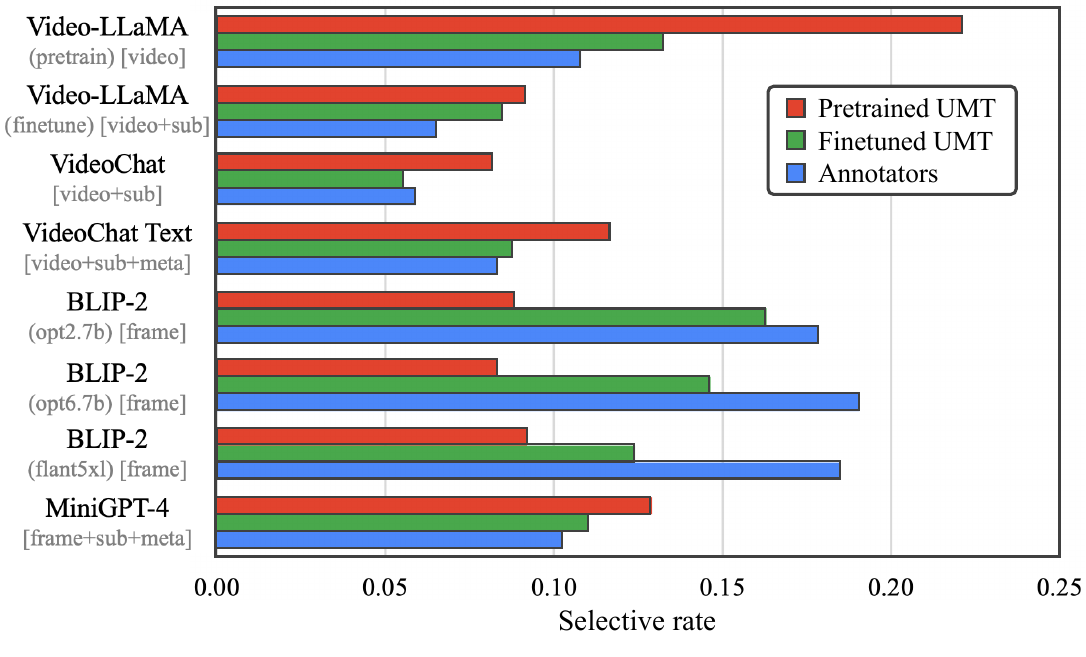}
    \vspace{-9mm}
    \mycaption{Distributions of the selective rate of teacher models}{
        We plot the distributions of the selective rate of eight teachers on 1,805 testing videos.
        The results are based on the selection of the pretrained ({\color{Red}red}) or finetuned ({\color{Green}green}) Unmasked Teacher~\cite{unmaskedteacher} and human annotators ({\color{NavyBlue}blue}).
    }
    \vspace{-6.5mm}
    \label{fig:teacher}
\end{figure}

\vspace{-1mm}
\subsection{Captioning with Cross-Modality Teachers}
\label{sec:teacher}
\vspace{-1.5mm}

Videos in HD-VILA-100M~\cite{hdvila} contain rich multimodal information beneficial for captioning.
Specifically, besides the video itself, there are also useful texts (\eg, video title, description, and subtitles) and images (\eg, individual video frames).
Driven by this insight, we propose to use several captioning models with the inputs of different modalities.

We start with a large pool including 31 captioning models.
The introduction of the model pool is in Appendix~\ref{app:31_teacher}.
Since running the inference of all models on 70M video clips is computationally expensive, we construct a short list of eight well-performing models based on a user study.
The list is shown in the y-axis of Figure~\ref{fig:teacher}.
More details of this process are in Appendix~\ref{app:user_study_all_good}.
%\sergey{which one?}
%
Briefly, the models are composed of five base models with different pretraining weights and input information.
The five base models include Video-LLaMA~\cite{videollama} (video VQA), VideoChat~\cite{videochat} (video VQA), VideoChat Text~\cite{videochat} (natural language model which textualizes the video content), BLIP-2~\cite{blipv2} (image captioning), and MiniGPT-4~\cite{minigpt4} (image VQA).
To implement video captioning by cross-modality teacher models, we formulate distinct captioning processes tailored to each modality.
For example, for the VQA models, in addition to visual data, we also input a prompt with additional text information and ask the models to summarize all multimodal inputs into one sentence.
%
%For the image-based models, 
% 
%With different combinations of input information and pre-trainining weights, we generate eight captions for each video.
%
Details on the captioning process of each teacher model are described in Appendix~\ref{app:teacher_inference}.
%\sergey{Which?}

% To prove the necessity of using multiple cross-modality teachers, we conduct a user study where annotators are asked to select the best caption from the 8 candidates. The task includes 2,000 videos (this is the testing set of the dataset introduced in Section~\ref{sec:v2t_retrieval}).
%
% We plot the distribution of the human selections in Figure~\ref{fig:teacher} ({\color{NavyBlue}blue} line).
%
% The results show that the best caption for different video is generated by often different teachers. The highest success rate of a single teacher (BLIP-2 with opt6.7b~\cite{opt}) is only $17.85\%$ implying the advantage of multiple models captioning
%
We hypothesize that teacher models using different modality data perform well on different kinds of videos.
For example, video models can perform better on videos with complex dynamics due to the additional modules to handle temporal information.
On the other hand, image models can accurately caption the videos with rare and uncommon objects, since they were trained using large-scale datasets of image-text pairs~\cite{laion}.
Finally, for videos that are visually hard to understand, VQA models have leverage as they can employ additional textual clues.

This hypothesis can be supported by a numerical evaluation.
Specifically, we conduct a user study where the participants are asked to select the best caption from eight candidates.
We plot the selective rate of each teacher model in Figure~\ref{fig:teacher} ({\color{NavyBlue}blue} bars).
The results show that the best captions are generated by different teacher models.
Moreover, the highest selective rate of an individual teacher model (\ie, BLIP-2 with opt6.7b~\cite{opt}) is only $17.85\%$. This fact expresses the limited captioning capability of a single model on a wide variety of videos.
% \sergey{how does this imply advantage of multimple teacher? To imply we need to show that all-bad rate?}.
% \tschen{have shown the all-bad rate in introduciotn section. I prefer not showing all-bad rate again here because these are from two user studys (select "all good" captions and select "the best" caption). I think showing two different all-bad rates is quite confusing.}

\vspace{-1.8mm}
\subsection{Fine-grained Video-to-Text Retrieval}
\label{sec:v2t_retrieval}
\vspace{-1.8mm}

Given multiple candidate captions for a video, we seek the one that best aligns with the video content.
An intuitive idea is to use the available generic video-to-text retrieval models~\cite{imagebind,unmaskedteacher} to pick such a caption. Unfortunately, we find that they usually fail to pick the optimal result.
One reason is that generic models are trained using contrastive learning objectives~\cite{simclr,coca} and learn to distinguish one sample from other completely unrelated samples\footnote{Negative samples for contrastive learning~\cite{simclr} are usually randomly sampled from the within-batch data and show no association to the anchor.}.
In contrast, in our case, all candidate captions are highly relevant to the video sample and require the model to discern subtle distinctions within each caption for optimal performance.

To tailor the retrieval model to our ``fine-grained'' retrieval scenario, we collect a subset of 100K videos, for which human annotators select the caption containing the most correct and detailed information about the main content of the video.
%
% We split this set to get 98K training and 2K testing samples.
%
We then finetune Unmasked Teacher~\cite{unmaskedteacher} (UMT) on this dataset.
We implement hard negative mining~\cite{albef,ifnd} for contrastive loss, where the seven captions not selected by annotators compose the hard negative samples and are assigned a larger training weight.
We describe the details of the dataset collection and finetuning of UMT in Appendix~\ref{app:retrieval_dataset} and~\ref{app:retrieval_finetune} respectively.

We quantitatively evaluate the retrieval performance of UMTs with and without finetuning on the validation set.
The experiments indicate that a finetuned UMT can achieve $35.90\%$ R@1 accuracy which significantly outperforms a pretrained UMT which has $21.82\%$ R@1.
Notably, we conducted a human agreement evaluation by asking two other persons to re-perform the annotation and comparing the results with the original annotations.
The average human agreement score is only $44.9\%$ R@1 showing that the task is subjective when more than one caption is equally good.
Alternatively, if we consider the captions selected by any of the three persons as good captions (\ie, a video might have multiple good captions), UMT achieves $78.9\%$ R@1.
Besides, in Figure~\ref{fig:teacher}, we show that a finetuned UMT ({\color{Green}green} bars) can select the captions distributed similarly to human-selected captions ({\color{NavyBlue}blue} bars).
We run the finetuned UMT on the whole dataset to select the best caption as the annotation as elaborated in Appendix~\ref{app:retrieval_inference}.

\begin{figure}[t]
    \centering
    \includegraphics[width=\linewidth]{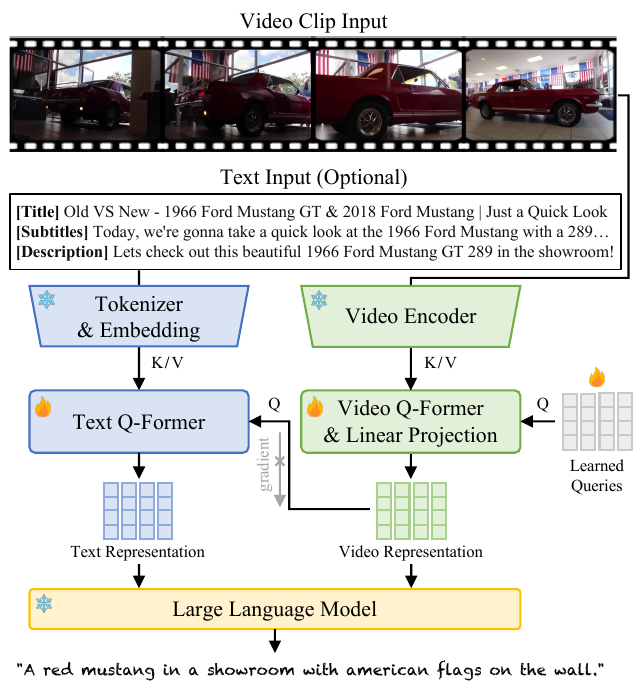}
    \vspace{-7mm}
    \mycaption{Architecture of student captioning model}{}
    \vspace{-4mm}
    \label{fig:architecture}
\end{figure}
\begin{table*}[t]
    \centering
    \small
    \mycaption{Zero-shot video captioning (\%)}{
        We compare Video-LLaMA~\cite{videollama} with official weight (pretrained on 2.5M videos and 595K images) and our Panda-2M pretraining weight.
        We also test our student model (with vision branch only) trained on the complete Panda-70M dataset.
        We report BLEU-4 (B-4)~\cite{bleu}, ROUGE-L (R)~\cite{rougel}, METEOR (M)~\cite{meteor}, CIDEr (C)~\cite{cider}, and BERTScore (BERT)~\cite{bertscore} on two benchmarks MSR-VTT~\cite{msrvtt} and MSVD~\cite{msvd}.
    }
    \label{tab:captioning}
    \vspace{-3mm}
    
    \setlength\tabcolsep{4.8pt}
    \begin{tabular}{lcccccccccccc}    
    \toprule
    \multirow{2}{*}{Method} & \multirow{2}{*}{Pretraining Data} &
    \multicolumn{5}{c}{MSR-VTT} & \multicolumn{5}{c}{MSVD} \\
    \cmidrule(lr){3-7} \cmidrule(lr){8-12}
    & & B4$\uparrow$ & R$\uparrow$ & M$\uparrow$ & C$\uparrow$ & BERT$\uparrow$ & B4$\uparrow$ & R$\uparrow$ & M$\uparrow$ & C$\uparrow$ & BERT$\uparrow$ \\
    \midrule
    %\textit{Zero-shot} &&&&&&&&&& \\
    Video-LLaMA~\cite{videollama} & 2.5M vid + 595K img &
    5.8 & 30.0 & 15.9 & 14.3 & 84.5 & 12.7 & 43.0 & 23.6 & 38.5 & 87.3 \\
    Video-LLaMA~\cite{videollama} & Panda-2M (Ours) &
    \underline{23.5} & \underline{48.6} & \underline{26.7} & \underline{29.1} & \underline{87.2} & \underline{31.2} & \underline{59.9} & \underline{34.7} & \underline{47.0} & \underline{89.8} \\
    \midrule
    \textbf{Student (Ours)} & \textbf{Panda-70M (Ours)} &
    \textbf{25.4} & \textbf{50.1} & \textbf{27.7} & \textbf{31.5} & \textbf{87.9} & \textbf{32.8} & \textbf{61.2} & \textbf{35.3} & \textbf{49.2} & \textbf{90.2} \\
    % \midrule
    % \midrule
    % \textit{Finetune} &&&&&&&&&& \\
    % Holder & - & 
    % 00.0 & 00.0 & 00.0 & 00.0 & 00.0 & 00.0 & 00.0 & 00.0 & 00.0 & 00.0 \\
    % \midrule
    % Video-LLaMA (vision)~\cite{videollama} & 2.5M videos + 3M images &
    % 00.0 & 00.0 & 00.0 & 00.0 & 00.0 & 48.4 & 74.7 & 41.1 & 113.3 & 91.6 \\
    % Video-LLaMA (vision)~\cite{videollama} & Panda-2M (Ours) &
    % 00.0 & 00.0 & 00.0 & 00.0 & 00.0 & 53.5 & 76.3 & 41.2 & 121.4 & 91.7 \\
    % Student & Panda-70M (Ours) &
    % 00.0 & 00.0 & 00.0 & 00.0 & 00.0 & 00.0 & 00.0 & 00.0 & 00.0 & 00.0 \\
    \bottomrule
    \end{tabular}
    \vspace{-2mm}
\end{table*}
\begin{table}[t]
    %\vspace{+1mm}
    \centering
    \small
    \mycaption{Comparison of the teacher(s) and student captioning models (\%)}{
        We conduct a user study to compare single teacher, all teacher, and two student models (with and without text).
        %We conduct an user study to ask participants to selects the best caption among ten options.
        %
        %Ten options are the predictions from eight teacher and two student models.
        %
        %Two students share the same model but with different inputs (with and without text).
        %
        %We also compare with the selective rate of all teacher models, which is the annotation of Panda-70M
    }
    \label{tab:userstudy}
    \vspace{-3mm}

    \setlength\tabcolsep{4.8pt}
    \begin{tabular}{lc}
    \toprule
    Model & Preference Ratio$\uparrow$ \\
    \midrule
    Video-LLaMA~\cite{videollama} (pretrain) &  9.4 \\
    Video-LLaMA~\cite{videollama} (finetune) &  7.0 \\
    VideoChat~\cite{videochat}               &  7.7 \\
    VideoChat Text~\cite{videochat}          &  3.3 \\
    BLIP-2~\cite{blipv2} (opt2.7b)           & 10.7 \\
    BLIP-2~\cite{blipv2} (opt6.7b)           &  9.0 \\
    BLIP-2~\cite{blipv2} (flant5xl)          &  9.9 \\
    MiniGPT-4~\cite{minigpt4}                &  3.1 \\
    \midrule
    Student (video input) (Ours)             & 18.4 \\
    Student (video+text inputs) (Ours)       & \underline{21.4} \\
    \midrule
    All Teachers (Ours)                      & \textbf{23.3} \\
    \bottomrule
    \end{tabular}
    \vspace{-2mm}
\end{table}

\vspace{-1mm}
\subsection{Multimodal Student Captioning Model}
\label{sec:student}
\vspace{-1.5mm}

While the aforementioned captioning pipeline can generate promising captions, the heavy computational demands hinder its capability to expand the dataset to an even larger scale. Indeed, one needs to run $8 + 1$ different models to annotate a single video clip. 
To deal with this problem, we learn a student captioning model on Panda-70M to distill the knowledge from multiple teacher models.

As shown in Figure~\ref{fig:architecture}, the student model includes visual and text branches, leveraging multimodal inputs.
For the vision branch, we use the same architecture as Video-LLaMA~\cite{videollama} to extract LLM-compatible video representation.
For the text branch, a straightforward design is to directly input text embedding into the LLM.
However, this will lead to two problems:
first, the text prompt with video description and subtitles can be too long, dominating the decision of the LLM and burdening heavy computation;
second, the information from the description and subtitles is often noisy and not necessary to align with the content of the video.
To tackle this, we add a text Q-former to extract the text representation with fixed length and better bridge the video and text representations. The Q-former has the same architecture as the Query Transformer in BLIP-2~\cite{blipv2}.
During training, we block the gradient propagation from the text branch to the vision branch and train the visual encoder only based on the video input.
More details about the architecture and training of the student model are in Appendix~\ref{app:student}.

% \sergey{did we ablate this? Does dimensionality reduction really help?}
% \tschen{yes, especially topic and subtitles are noisy and not well-aligned. directly inputting will let LLM just output video topic or subtitles}
% \tschen{another concerns is computation complexity. Video embedding has length of 32; the same length for text is just like 20~30 words which cannot include all of topic/subtitles/description. That's why we want to do dimensionality reduction.}

\newcommand{\nor}{{} {} - {} {} {} {}}
\begin{table*}[t]
    \centering
    \footnotesize
    \mycaption{Video and text retrieval (\%)}{
        We compare the Unmasked Teacher~\cite{unmaskedteacher} with the official checkpoint (pretrained on 2.5M videos and 3M images) and our Panda-5M pretraining.
        We evaluate their performance on zero-shot and finetune text-to-video (T2V) and video-to-text (V2T) retrieval.
        We report R@1, R@5, and R@10 accuracy on three benchmarks: MSR-VTT~\cite{msrvtt}, DiDeMo~\cite{didemo}, and MSVD~\cite{msvd}.
    }
    \label{tab:retrieval}
    \vspace{-3mm}
    
    \setlength\tabcolsep{2.5pt}
    \begin{tabular}{lcccccccccc}    
    \toprule
    \multirow{2}{*}{Method} & \multirow{2}{*}{Pretraining Data} &
    \multicolumn{3}{c}{MSR-VTT} & \multicolumn{3}{c}{DiDeMo} & \multicolumn{3}{c}{MSVD} \\
    \cmidrule(lr){3-5} \cmidrule(lr){6-8} \cmidrule(lr){9-11}
    & & R@1$\uparrow$ & R@5$\uparrow$ & R@10$\uparrow$ &
        R@1$\uparrow$ & R@5$\uparrow$ & R@10$\uparrow$ &
        R@1$\uparrow$ & R@5$\uparrow$ & R@10$\uparrow$\\
    \midrule
    \midrule
    \multicolumn{2}{l}{\textit{Zero-shot T2V / V2T Retrieval}} &&&&&&&&& \\
    AlignPrompt~\cite{alignprompt}      & 2.5M vid + 3M img &
    24.1 / \nor & 44.7 / \nor & 55.4 / \nor &
    23.8 / \nor & 47.3 / \nor & 57.9 / \nor &
    \nor / \nor & \nor / \nor & \nor / \nor \\
    BridgeFormer~\cite{bridgeformer}    & 2.5M vid + 3M img &
    26.0 / \nor & 46.4 / \nor & 56.4 / \nor &
    25.6 / \nor & 50.6 / \nor & 61.6 / \nor &
    43.6 / \nor & 74.9 / \nor & 84.9 / \nor \\
    \midrule
    UMT~\cite{unmaskedteacher}          & 2.5M vid + 3M img & 
    \underline{30.2} / \underline{33.3} & \underline{51.3} / \underline{58.1} & \underline{61.6} / \underline{66.7} &
    \underline{33.6} / \underline{32.1} & \underline{58.1} / \underline{57.3} & \underline{65.5} / \textbf{66.7} &
    \underline{66.3} / \textbf{44.4}    & \underline{85.5} / \textbf{73.3}    & \underline{89.3} / \textbf{82.4} \\
    UMT~\cite{unmaskedteacher}          & \textbf{Panda-5M (Ours)} & 
    \textbf{37.2}    / \textbf{36.3}    & \textbf{58.1}    / \textbf{61.0}    & \textbf{69.5}    / \textbf{69.7} &
    \textbf{34.2}    / \textbf{33.4}    & \textbf{58.4}    / \textbf{57.9}    & \textbf{66.5}    / \underline{65.8} &
    \textbf{71.2}    / \underline{37.2} & \textbf{88.4}    / \underline{65.1} & \textbf{92.7}    / \underline{75.6} \\
    \midrule
    \midrule
    \multicolumn{2}{l}{\textit{Finetune T2V / V2T Retrieval}} &&&&&&&& \\
    CLIP4Clip~\cite{clip4clip}          & 400M img & 
    44.5 / 40.6 & 71.4 / 69.5 & 81.6 / 79.5 &
    43.4 / 42.5 & 70.2 / 70.6 & 80.6 / 80.2 &
    46.2 / 62.0 & 76.1 / 87.3 & 84.6 / 92.6 \\
    X-CLIP~\cite{xclip}                 & 400M img &
    49.3 / 48.9 & 75.8 / \underline{76.8} & \underline{84.8} / \underline{84.5} &
    50.4 / \textbf{66.8} & 80.6 / \textbf{90.4} & \nor / \nor &
    47.8 / 47.8 & 79.3 / 76.8 & \nor / \nor \\
    InternVideo~\cite{internvideo}      & 146M vid + 100M img &
    \underline{55.2} / \underline{57.9} & \nor / \nor & \nor / \nor &
    57.9 / 59.1 & \nor / \nor & \nor / \nor &
    \textbf{58.4} / 76.3 & \nor / \nor & \nor / \nor \\
    \midrule
    UMT~\cite{unmaskedteacher}          & 2.5M vid + 3M img & 
    53.3             / 51.4             & \underline{76.6} / 76.3             & 83.9             / 82.8 &
    \underline{59.7} / \underline{59.5} & \underline{84.9} / 84.5             & \underline{90.8} / \textbf{90.7} &
    53.7             / \underline{77.2} & \underline{80.5} / \underline{91.6} & \underline{86.8} / \underline{94.8} \\
    UMT~\cite{unmaskedteacher}          & \textbf{Panda-5M (Ours)} & 
    \textbf{58.4}    / \textbf{58.5}    & \textbf{80.9}    / \textbf{81.0}    & \textbf{86.9}    / \textbf{87.0} &
    \textbf{60.6}    / 58.9             & \textbf{86.0}    / \underline{84.6} & \textbf{92.4}    / \underline{90.4} &
    \underline{57.5} / \textbf{81.3}    & \textbf{83.6}    / \textbf{93.7}    & \textbf{89.5}    / \textbf{96.6} \\
    \bottomrule
    \end{tabular}
    \vspace{-1.5mm}

\end{table*}

\section{Experiments}
\label{sec:experiments}
\vspace{-1.5mm}

We visualize the samples of Panda-70M in Appendix~\ref{app:visualization}.
To quantitatively evaluate the effectiveness of Panda-70M, we test its pretraining performance on three downstream applications: video captioning in Section~\ref{sec:captioning}, video and text retrieval in Section~\ref{sec:retrieval}, and video generation in Section~\ref{sec:generation}.
%
% The training details can be found in Appendix~\ref{app:training_details}.
The training details of the downstream models adhere to the official codebases unless explicitly specified.

\vspace{-1mm}
\subsection{Video Captioning}
\label{sec:captioning}
\vspace{-1.5mm}

\noindent \textbf{Experiment setup.}
To evaluate the performance of video captioning, we use Video-LLaMA~\cite{videollama} with the vision branch only as the base model.
We compare two pretraining weights: the official weight, which is jointly trained on 2.5M video-text pairs and 595K image-text pairs~\cite{llava}, and the weight trained on our Panda-2M from scratch.
Panda-2M is a randomly sampled subset of Panda-70M and shares the same amount of training samples as the official weight.
We also train our student model with both video and text branches on complete Panda-70M for better captioning performance.
For all models, we use the same backbone, using Vicuna-7B~\cite{vicuna} as the large-language model, ViT~\cite{eva} and Q-Former~\cite{blipv2} as the video encoder, and the linear projection layer from MiniGPT-4~\cite{minigpt4}.
For Panda-2M pretraining, we only use the video and caption data without using other textual information for a fair comparison.
%
% We set a batch size of 64 and train the model for 800K iterations.
%
For the student model, in addition to the video, we also randomly input the metadata and subtitles into the model during training.
%
%More training details can be found in the Appendix.
%
%We last the training for 2M training steps with a batch size of 48.
%

\begin{figure}[t]
    \centering
    \includegraphics[width=\linewidth]{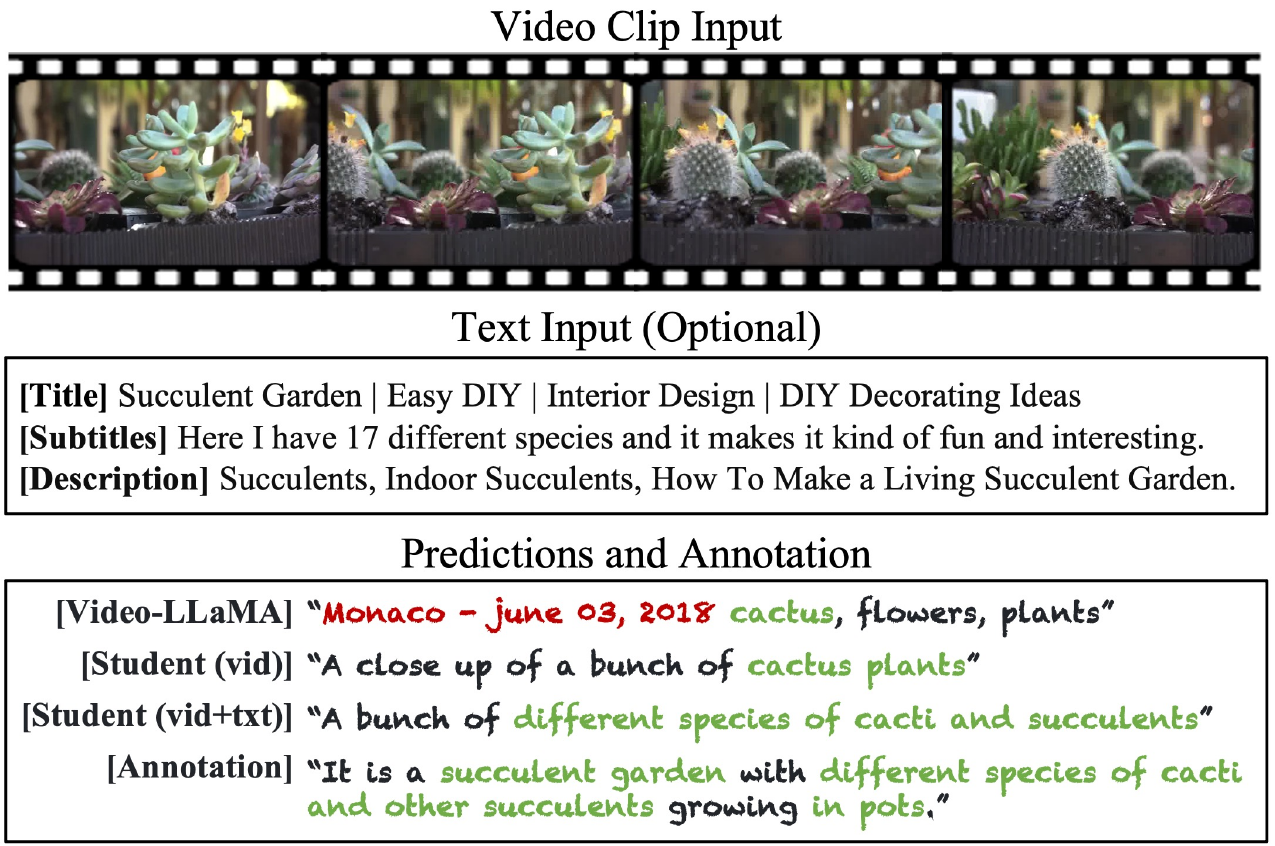} 
    \vspace{-6mm}
    \mycaption{Qualitative comparison of video captioning}{
        We visualize a sample from the testing set of Panda-70M and show its annotation (bottommost).
        We also show the captions predicted from three models, including Video-LLaMA~\cite{videollama} with official weight and the student models with video-only or video and text inputs.
    }
    \vspace{-5mm}
    \label{fig:captioning}
\end{figure}

\noindent \textbf{Downstream datasets and evaluation metrics.}
We test zero-shot video captioning on two benchmarks: MSR-VTT~\cite{msrvtt} and MSVD~\cite{msvd}.
MSR-VTT contains 10K videos with 20 manually annotated captions for each video; we report the results on the 2,990 testing split.
MSVD consists of 1,970 videos with a total of 80K descriptions; we report the numbers on the 670 testing videos.
Note that we do not use any training or validation videos from the downstream datasets.
To quantitatively evaluate the quality of output captions, we follow the common protocols~\cite{univl,swinbert,mplug2} and report BLEU-4~\cite{bleu}, ROGUE-L~\cite{rougel}, METEOR~\cite{meteor}, and CIDEr~\cite{cider}.
All the metrics are computed using the pycocoevalcap~\cite{coco} package.
We also compute BERTScore~\cite{bertscore} to evaluate the contextual similarity for each token in the ground truth and the predicted captions.
The results are reported in Table~\ref{tab:captioning}.
For a fair comparison, we do not input any additional text information to the student model during the inference on the downstream datasets.
In Figure~\ref{fig:captioning}, we also showcase a video sample from the testing set of Panda-70M and the predicted captions for the qualitative comparison.

As in Table~\ref{tab:captioning}, Video-LLaMA with Panda-2M pretraining weight achieves significantly superior performance compared to the official weight.
Numerically, our pretraining weight yields $17.7\%$ and $18.5\%$ improvement respectively on MSR-VTT and MSVD in terms of B-4.
Besides, in Figure~\ref{fig:captioning}, we can find that the caption from the original Video-LLaMA contains irrelevant and generic information, such as date and location.
In comparison, our prediction better aligns with the video content.

\noindent \textbf{Can the student perform better than its teacher?}
In Section~\ref{sec:student}, we learn a student model in a knowledge distillation manner.
To evaluate the performance of the student model, we conduct a user study where participants are asked to select the best caption from ten candidates for each video.
Ten captions are predicted from eight teacher models and two student models (with and without text inputs).
We collect the results from five participants to reduce the personal subjective bias.
Each participant saw the same 200 videos, which were randomly sampled from the testing set and had not been seen during the training of the student model and UMT.
We report the preference ratio of each model and the R@1 accuracy of the finetuned UMT (\ie, all teachers) in Table~\ref{tab:userstudy}.
%
%More details about the user study are in Appendix~\ref{app:user_evaluation}.
%
We can observe that the student model outperforms any individual teacher model and achieves a comparable performance with all teacher models.

\noindent \textbf{Can multimodal inputs leverage video captioning?}
Our student model supports both video and text inputs.
In Table~\ref{tab:userstudy}, we show that the student model with both video and text inputs outperforms the model with video input only by $3.0\%$ preference ratio.
Qualitatively, we show the predictions with and without text inputs in Figure~\ref{fig:captioning}.
While the prediction with pure video input can include partial content of the video, like ``cactus'', the model with both video and text inputs can more comprehensively include keywords such as ``succulents'' and ``different species'' from the video title, description, and subtitles.

\vspace{-1mm}
\subsection{Video and Text Retrieval}
\label{sec:retrieval}
\vspace{-1mm}

\noindent \textbf{Experiment setup.}
We use Unmasked Teacher~\cite{unmaskedteacher} as the base model to evaluate the performance on video and text retrieval.
The standard protocols~\cite{alignprompt,bridgeformer,unmaskedteacher,vindlu,mplug2} jointly use 3M images from CC3M~\cite{cc3m} and 2.5M videos as the pretraining datasets.
Thus, we randomly sample a Panda-5M subset, which shares the same number of training samples as the standard pretraining dataset for a fair comparison.
%Panda-5M is a 5.5M subset randomly sampled from Panda-70M. 
%
%We compare the pretraining performance of these two datasets.
%
For both datasets, we use the same backbone composed of ViT-L/16~\cite{transformers} and BERTlarge~\cite{bert}.
We use the official weights for the standard datasets pretraining and train the model from scratch for our Panda-5M.
%
%Specifically, following the official codebase, we initialize the ViT with Kinetic-710 pretraining weight~\cite{uniformerv2} and train the whole model using CLIP-ViTL/14~\cite{clip} as teachers. 
%
%We sample 4 frames~\cite{singularity} for each video sample and train for 10 epochs.

\noindent \textbf{Downstream datasets and evaluation metric.}
We test both zero-shot and finetune retrieval on three benchmarks: MSR-VTT~\cite{msrvtt}, DiDeMo~\cite{didemo}, and MSVD~\cite{msvd}.
For MSR-VTT, we follow the common protocol~\cite{jsfusion,alignprompt} to evaluate on 1K testing split, which is not the same as the testing videos for captioning in Section~\ref{sec:captioning}.
For DiDeMo~\cite{didemo}, it contains 10K Flickr videos with a total of 40K dense captions.
As in the previous standard~\cite{ce,clipbert,bridgeformer}, we evaluate paragraph-to-video retrieval by concatenating all sentence descriptions of one video into a single query.
We report the results on the 1K testing set.
For MSVD~\cite{msvd}, we report the results on the 670 testing videos.
%
%We do not use any training videos from the downstream datasets and none of the testing videos and the text prompts have been seen during training.
%
We employ the standard metric and report R@1, R@5, and R@10 accuracy on both text-to-video and video-to-text retrieval in Table~\ref{tab:retrieval}. 
%the rank-1 (R@1), rank-5 (R@5), and rank-10 (R@10) accuracy in cumulative matching curve (CMC) 

%\noindent \textbf{Quantitative Results.}
We can observe that pretraining with our Panda-5M outperforms the official weight in both zero-shot and finetune retrieval settings.
Especially, our pretraining yields \textbf{$7.0\%$}, \textbf{$0.6\%$}, and \textbf{$4.9\%$} lifts in terms of R@1 of zero-shot text-to-video retrieval on MSR-VTT~\cite{msrvtt}, DiDeMo~\cite{didemo}, and MSVD~\cite{msvd} respectively.
Besides, pretraining UMT~\cite{unmaskedteacher} with our Panda-5M also outperforms the existing state-of-the-art methods~\cite{clip4clip, xclip, internvideo} which are pretrained with much more vision-text data pairs (\ie, $>$100M).

\vspace{-1mm}
\subsection{Text-to-Video Generation}
\label{sec:generation}
\vspace{-1mm}

\noindent \textbf{Experiment setup.}
To evaluate the effectiveness of text-to-video generation, we use AnimateDiff~\cite{animatediff} as the base model and compare two weights: the officially released weight, which is trained on 2.5M text-video pairs, and the weight trained on our Panda-2M, a 2.5M subset of Panda-70M.
We follow the official codebase and use Stable Diffusion v1.5~\cite{ldm} (SD) as the base text-to-image (T2I) generator.
During training, we fix T2I modules and only train the motion modeling modules.
For each training video, we sample 16 frames with a stride of 4, and then resize and center-crop to $256 \times 256\mathrm{px}$ resolution.
%
%We set the batch size of 4 and last the training for 950K steps.
%steps and use the AdamW~\cite{adamw} optimizer with a learning rate of $1e^{-4}$, $\beta=[0.9, 0.999]$, and a weight decay of 0.01.

\begin{table}[t]
    \centering
    \small
    \mycaption{Zero-shot text-to-video generation}{
        We compare the zero-shot text-to-video generation of AnimateDiff~\cite{animatediff} with the official weight (pretrained on 2.5 M videos) and our Panda-2M pretraining.
        We report FVD~\cite{fvd} on UCF101~\cite{ucf101} and CLIP similarity (CLIPSim)~\cite{clipsim} on MSR-VTT~\cite{msrvtt}.
        We only compare with the models trained with less than 10M videos.
        % CogVideo: no mention
        % MagicVideo (10M): subsets of HD-VILA-100M 
        % LVDM (18K): UCF-101 (13,320), TaiChi (3,334), Sky Time-lapse (1,167)
        % Modelscope: LAION2B
        % VideoFactory: HD-VG-130M
        %
    }
    \label{tab:generation}

    \vspace{-3mm}
    \setlength\tabcolsep{4.8pt}
    \begin{tabular}{lccc}
    \toprule
    \multirow{2}{*}{Method} & \multirow{2}{*}{(\#) P-T Videos}
    & UCF101          & MSR-VTT \\
    \cmidrule(lr){3-3} \cmidrule(lr){4-4}
    & & FVD$\downarrow$ & CLIPSim$\uparrow$ \\
    \midrule
    CogVideo~\cite{cogvideo}     & 5M   & 701.6 & -      \\
    MagicVideo~\cite{magicvideo} & 10M  & 699.0 & -      \\
    LVDM~\cite{lvdm}             & 18K  & 641.8 & 0.2751 \\
    ModelScope~\cite{modelscope} & 10M  & 639.9 & \textbf{0.3000} \\
    VideoLDM~\cite{videoldm}     & 10M  & 550.6 & -      \\
    %VideoFactory                 & 140M & 410.0 & 0.3070 \\
    \midrule
    AnimateDiff~\cite{animatediff} & 2.5M & \underline{499.3} & 0.2869 \\
    AnimateDiff~\cite{animatediff} & \textbf{Panda2M (Ours)} & \textbf{421.9} & \underline{0.2880} \\
    \bottomrule
    \end{tabular}
    \vspace{-3mm}
\end{table}

\begin{figure}[t]
    \centering
    \includegraphics[width=\linewidth]{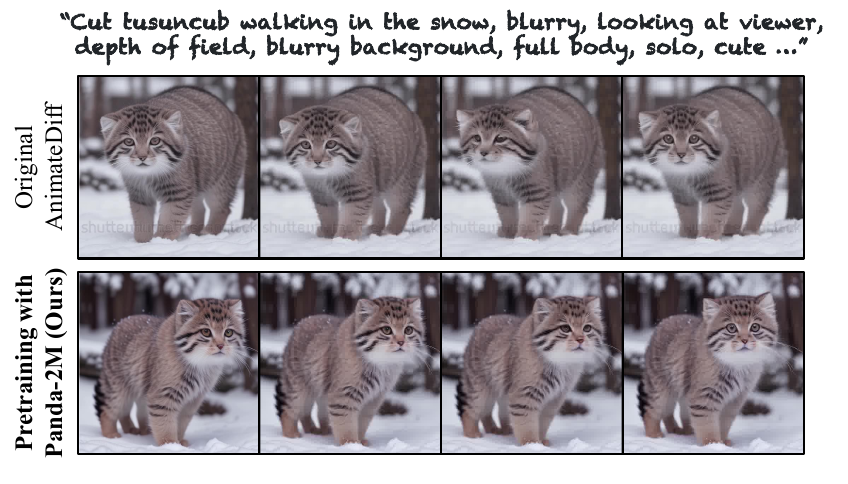}
    \vspace{-8mm}
    \mycaption{Qualitative results of text-to-video generation}{
        We visualize the videos generated by the AnimateDiff~\cite{animatediff} with official weight (top) and our Panda-2M pretraining (bottom).
        Note that the test prompt and the video sample of the original AnimateDiff (top) are directly from the project website of AnimateDiff.
    }
    \vspace{-3mm}
    \label{fig:generation}
\end{figure}

\noindent \textbf{Downstream datasets and evaluation metrics.}
To evaluate the models, we follow the evaluation protocols~\cite{videoldm,videofactory,makeavideo,pyoco} for zero-shot evaluation on UCF101~\cite{ucf101} and MSR-VTT~\cite{msrvtt}.
Specifically, we generate 16-frame videos in $256\times256 \mathrm{px}$ resolution.
For UCF101~\cite{ucf101}, we produce a text prompt for each class~\cite{pyoco} and generate 10,000 videos which share the same class distribution as the original dataset~\cite{videoldm,videofactory}.
We compute Fréchet Video Distance (FVD)~\cite{fvd} on the I3D embeddings~\cite{digan}.
For MSR-VTT~\cite{msrvtt}, we generate a video sample for each of the 59,800 test prompts~\cite{makeavideo,pyoco} and compute CLIP similarity (CLIPSim)~\cite{clipsim}.
We report the numbers in Table~\ref{tab:generation}.
We also show the generated video samples in Figure~\ref{fig:generation}.
To visualize the results, we follow the official codebase and replace SD T2I with personalized Dreambooth weight~\cite{dreambooth}, TUSUN\footnote{https://civitai.com/models/33194/pallass-catmanul-lora}.
Note that the test prompt and the video sample from the AnimtateDiff with the official weight (top row in Figure~\ref{fig:generation}) are directly from the project page of AnimateDiff.

%\noindent \textbf{Quantitative and Qualitative Results.}
Panda-2M pretraining consistently shows superior performance on both metrics compared to the official weight.
As highlighted, our pretraining yields $77.4$ lower FVD on UCF101 and outperforms state-of-the-art models pretrained on a dataset within a 10M scale in terms of FVD.
Qualitatively, our pretraining weights can generate the video with a more meaningful motion and photorealistic appearance and do not include a watermark.

\vspace{-2mm}
\section{Conclusion and Limitations}
\label{sec:conclusion}
\vspace{-1.5mm}

This paper introduces Panda-70M, a large-scale video dataset with caption annotations.
The dataset includes high-resolution and semantically coherent video samples.
To caption 70M videos, we propose an automatic pipeline that can leverage multimodal information, such as video description, subtitles, and individual static video frames.
We demonstrate that pretraining with Panda-70M can facilitate three downstream tasks: video captioning, video and text retrieval, and text-to-video generation.

Despite showing impressive results, the proposed dataset is still bound by a few limitations. 
First, we collect the videos from HD-VILA-100M~\cite{hdvila}, where most of the samples are vocal-intensive videos.
Hence, the major categories of our dataset are news, television shows, documentary films, egocentric videos, and instructional and narrative videos.
As our annotation pipeline does not require the presence of video subtitles, we list the collection of more unvocal videos as an important extension of this work.

Second, we focus on a fine-grained dataset where the video samples are semantically consistent so the caption can accurately express its semantics content without ambiguity.
Nevertheless, it would limit the content diversity within a single video and also reduce average video duration, which might be hurtful to the downstream tasks, such as long video generation~\cite{longvideo} and dense video captioning~\cite{pdvc,vid2seq}.
Future efforts in building datasets with long videos and dense captions can benefit these downstream applications.

\vspace{+1mm}
\noindent \textbf{Risk mitigation.}
Prior to the release of the dataset, we used the internal automatic pipeline to filter out the video samples with harmful or violent language and texts that include drugs or hateful speech. We also use the NLTK framework to replace all people's names with ``person''. 
\vspace{-5mm}
{
    \small
    \bibliographystyle{ieeenat_fullname}
    \bibliography{main}
}

\clearpage
\setcounter{page}{1}
\appendix

\begin{nolinenumbers}
\maketitlesupplementary
\vspace{-3.0em}
\addcontentsline{toc}{section}{Appendix} % Add the appendix text to the document TOC

\part{} % Start the appendix part
\parttoc % Insert the appendix TOC
\end{nolinenumbers}

\section{Details of Semantics-Aware Video Splitting Algorithm}
\label{app:splitting}
In Section~\ref{sec:splitting}, we propose a video splitting algorithm to cut a long video into several semantically coherent clips.
The algorithm includes two stages, splitting and stitching, for which the details are described in Appendix~\ref{app:stage1_splitting} and~\ref{app:stage2_stitching}.

\subsection{Stage1: Splitting based on Shot Boundary Detection}
\label{app:stage1_splitting}
We first split a long video by PySceneDetect~\cite{pyscenedetect}.
Specifically, we use $\mathrm{ContentDetector}$ with $\mathrm{cutscene\_threshold}$ of 25 and $\mathrm{min\_scene\_len}$ of 15 frames.
Next, we design a two-step post-processing algorithm to handle 1) long videos with complex transitions, such as fade-in and fade-out effects, that cannot be reliably detected by PySceneDetect and 2) unedited footage that does not contain any cut-scenes but has semantic changes within the same clip.

To handle both cases, we propose creating artificial scene cuts each 5 seconds for clips without cut-scene.
That is, if a video clip is longer than 5 seconds, we cut out the first 5 seconds as a new clip and recursively apply the same procedure to the remaining part.
%
% We observe that most of the short video clips without cut scene is semantically consistent.
%
Since we are only interested in semantically consistent video clips, we extract the ImageBind~\cite{imagebind} features of the frames near the beginning or the end.
If the features of these two frames are dramatically different we remove that clip.
Specifically, given a $n$-frame video clip $C$, we extract the features $f(C_A)$ and $f(C_B)$ for the number $0.1 \times n$ and $0.9 \times n$ frames, denoted as $C_A$ and $C_B$.
We only keep the video clips if satisfying $\lVert f(C_A) - f(C_B) \rVert \leq 1.0$.
As such, we can exclude video clips with transition effects or significant semantics changes within a clip.

\subsection{Stage2: Stitching based on Semantics Similarity}
\label{app:stage2_stitching}
The first stage introduces many short consecutive clips with the same semantic content. 
To this end, we propose an additional procedure to merge the clips with the same semantic content.
%To avoid splitting a long video into lots of fragmentary clips, we then stitch the adjacent video clips into one if they are semantically similar.
%
Formally, given two adjacent clips $C^1$ and $C^2$ in sequence, we concatenate them into a clip if $\lVert f(C^1_B) - f(C^2_A) \rVert \leq 0.6$.

Finally, we perform a post-processing to stabilize the quality and diversity of the video clips with the following steps:

\begin{compactitem}
\item First, we exclude the clips shorter than $2$ seconds or clips that contain only slight motion (\ie, $\lVert f(C_A) - f(C_B) \rVert \leq 0.15$). For the videos longer than $60$ seconds, we only retain the first $60$ seconds.
\item Next, we represent each clip by the average of ImageBind features extracted from stage1 (Section \ref{app:stage1_splitting}) and only keep the video clips that are semantically different (\ie, Euclidean distance $> 0.3$) from the precedent clips to increase the diversity of the video samples.
\item Finally, we trim out the first and last $10\%$ of a video clip as we notice that the beginning and the ending of a clip usually contain unstable camera movement or transition effects.
\end{compactitem}

\begin{figure}[t]
    \centering
    \includegraphics[width=\linewidth]{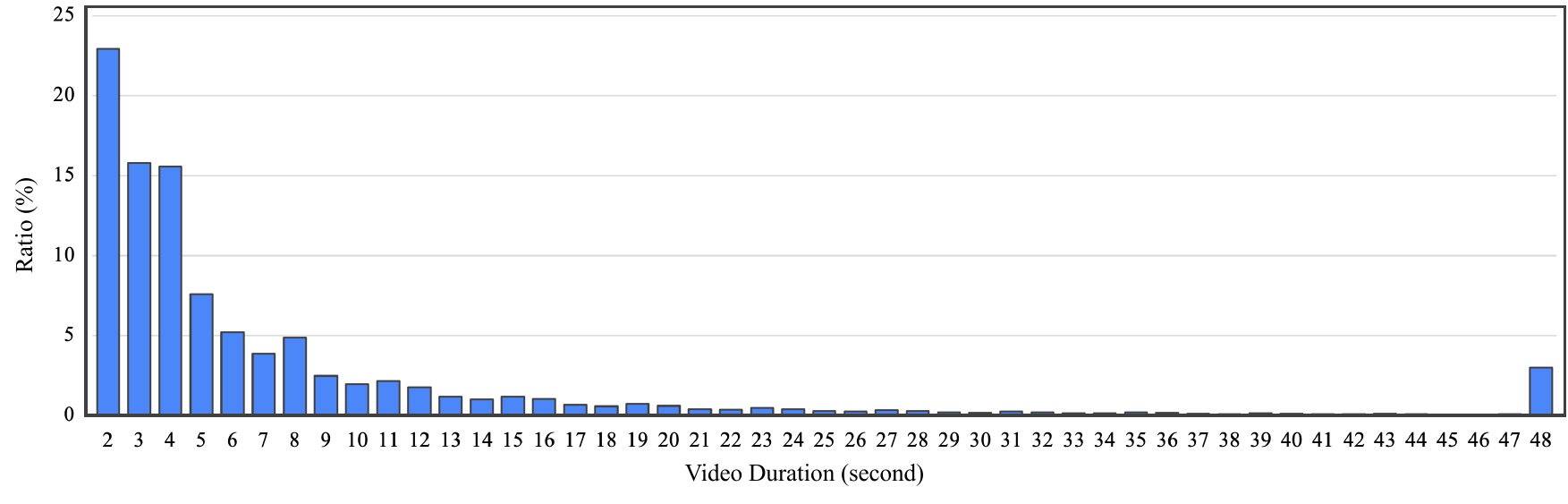}
    \vspace{-7mm}
    \mycaption{Distribution of video duration of Panda-70M}{}
    %\vspace{-3mm}
    \label{fig:duration}
\end{figure}

With the proposed splitting algorithm, we split $3,790,459$ long videos into $70,817,169$ clips with an average clip duration of $8.477$ seconds.
We plot the distribution of video length in Figure~\ref{fig:duration}.

\section{Details of Teacher Captioning Models: Pool, Inference, and Selection}
\label{app:teacher}
In Section~\ref{sec:teacher}, we propose to use multiple cross-modality teacher models for captioning.
Specifically, we start with a large pool including 31 captioning models.
We elaborate on the composition of the model pool and how we implement them for video captioning in Appendix~\ref{app:31_teacher} and~\ref{app:teacher_inference} respectively.
As running the inference of the models to 70M videos is computationally expensive, we select only 8 models as the representative, based on a human evaluation.
We will describe more details about this process in Appendix~\ref{app:user_study_all_good}.

\begin{table*}[t]
    \centering
    \small
    \mycaption{Overview of 31 teacher models}{
        31 teacher models are composed of 6 base models with various weights and input information.
        Input data includes vision (V), subtitles (S), and metadata (M).
        Vision data is either a video or a static video frame, depending on the type of base model.
        Metadata includes the title and the description of a video.
        For example, V-S-M for MiniGPT-4 means MiniGPT-4 with the inputs of a video frame, subtitles, and metadata.
    }
    \label{tab:teachers}
    \vspace{-2mm}
    
    \setlength\tabcolsep{4.8pt}
    \begin{tabular}{lccccccc}    
    \toprule
    \multirow{2}{*}{Base Model} & \multirow{2}{*}{Type}& \multirow{2}{*}{Weights} & \multicolumn{4}{c}{Input Information} & \multirow{2}{*}{\# of Models} \\
    \cmidrule(lr){4-7}
    & & & V & V-S & V-M & V-S-M \\
    \midrule
    Video-LLaMA~\cite{videollama}      & Video VQA           & pretrain / finetune & \cmark & \cmark & \cmark & \cmark & 8 \\
    VideoChat~\cite{videochat}         & Video VQA           & 7B                  & \cmark & \cmark & \cmark & \cmark & 4 \\
    VideoChat Text~\cite{videochat}    & NLP-based Video VQA & -                   & \cmark & \cmark & \cmark & \cmark & 4 \\
    Video-ChatGPT~\cite{video_chatgpt} & Video VQA           & -                   & \cmark & \cmark & \cmark & \cmark & 4 \\
    BLIP-2~\cite{blipv2}               & Image Captioning    & opt2.7b / opt6.7b / flant5xl & \cmark & \xmark & \xmark & \xmark & 3 \\
    MiniGPT-4~\cite{minigpt4}          & Image VQA           & 7B / 13B            & \cmark & \cmark & \cmark & \cmark & 8 \\
    \bottomrule
    \end{tabular}
    \vspace{-1mm}
\end{table*}

\subsection{Introduction of 31 Captioning Models Pool}
\label{app:31_teacher}
The primary reason to utilize cross-modality teacher models is to leverage multimodal data that would benefit video captioning.
As such, we consider the base models, including image/video visual-question-answering (VQA) and image captioning models.
Specifically, we employ Video-LLaMA~\cite{videollama}, VideoChat~\cite{videochat}, VideoChat Text~\cite{videochat}, Video-ChatGPT~\cite{video_chatgpt}, BLIP-2~\cite{blipv2}, and MiniGPT-4~\cite{minigpt4} as the base models.
Based on these models, we collect 31 captioning models in total using different weights and input information.
We list the summary of all captioning models in Table~\ref{tab:teachers}.

% For videochat-text we use vicuna instead of chatgpt

\subsection{Inference of Cross-Modality Teacher Model for Video Captioning}
\label{app:teacher_inference}
We list the inference details of each base model as follows:
\begin{compactitem}
\item \textbf{Video-LLaMA}~\cite{videollama}
is a video VQA model. We only use the vision branch and do not use the audio one.
The model uses Vicuna-7B~\cite{vicuna} as the LLM to implement VQA.
We use two official weights, including the pretraining weight, which is trained on 2.5M video-text pairs and LLaVA-CC3M~\cite{llava}, and the finetuning weight, which is further finetuned on instruction-tuning data from~\cite{minigpt4,llava,videochat}.
\item \textbf{VideoChat}~\cite{videochat} \textbf{and Video-ChatGPT}~\cite{video_chatgpt} 
are video VQA models. We use Vicuna-7B as the LLM and follow the official codebase for the rest of the configuration.
\item \textbf{VideoChat Text}~\cite{videochat}
is a natural-language processing (NLP)-based video VQA model.
The model would textualize the video content into video tags, dense captions, and a general caption respectively by three models~\cite{swintransformer,grit,t5}.
As such, users can have a conversation with a chatbot and discuss the video based on the extracted textual content.
The original codebase uses ChatGPT-4~\cite{chatgpt4} as the chatbot, which is, however, not freely released to the public.
Thus, we replace it with LLaMA~\cite{llama} for large-scale captioning.
\item \textbf{BLIP-2}~\cite{blipv2}
is a language-image pretraining model. We only use it for image captioning and do not input texts.
We use the weights pretraining with different LLMs, including OPT~\cite{opt} (opt2.7b and opt6.7b) and FlanT5~\cite{flant5} (flant5xl).
\item \textbf{MiniGPT-4}~\cite{minigpt4}
is an image VQA model. We use two variants respectively with Vicuna-7B and Vicuna-13B as LLMs.
\end{compactitem}

\begin{figure}[t]
    \centering
    \includegraphics[width=.9\linewidth]{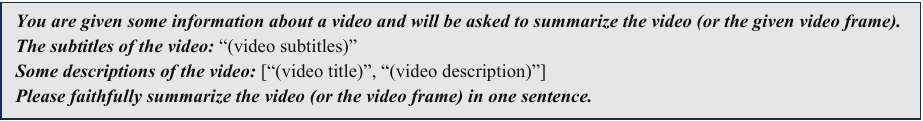}
    %\vspace{-1mm}
    \mycaption{Prompt template of the VQA models}{}
    %\vspace{-1mm}
    \label{fig:prompt}
\end{figure}

\vspace{+1mm}
To implement cross-modality teacher models for video captioning, we design the algorithms specifically for the models of different modalities.
For an image model, given an $n$-frame video clip, we randomly sample a video frame in-between number $0.3 \times N$ and $0.7 \times N$ frames as the input.
For a VQA model, in addition to the visual data, we also input a text prompt that could include additional textual information, such as video title, description, and subtitles, to assist video captioning.
Specifically, we use the prompt template in Figure~\ref{fig:prompt} if we would like to include the information of either metadata or subtitles or both for captioning.
In contrast, we use a dummy prompt: ``\textit{Please faithfully summarize the video (or image) in one sentence.}'' if we only input the vision data for captioning.

\begin{figure}[t]
    \centering
    \includegraphics[width=.8\linewidth]{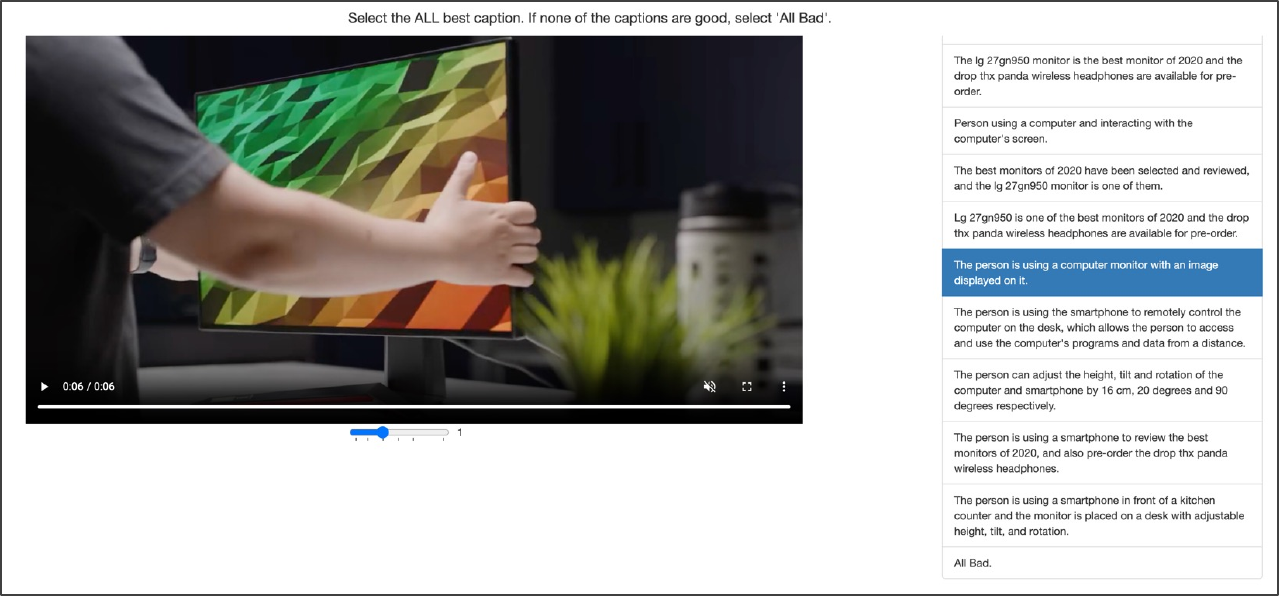}
    \vspace{-1mm}
    \mycaption{Screenshot of the user study interface}{}
    \label{fig:interface}
\end{figure}

\subsection{Selecting 8 Captioning Models based on a Human Evaluation}
\label{app:user_study_all_good}
Running 31 captioning models on 70M videos requires significant computation resources.
Hence, we propose to find a well-performing subset of the models by a two-step algorithm, including a human evaluation and model selection algorithm.

\begin{figure}[t]
    \centering
    \includegraphics[width=\linewidth]{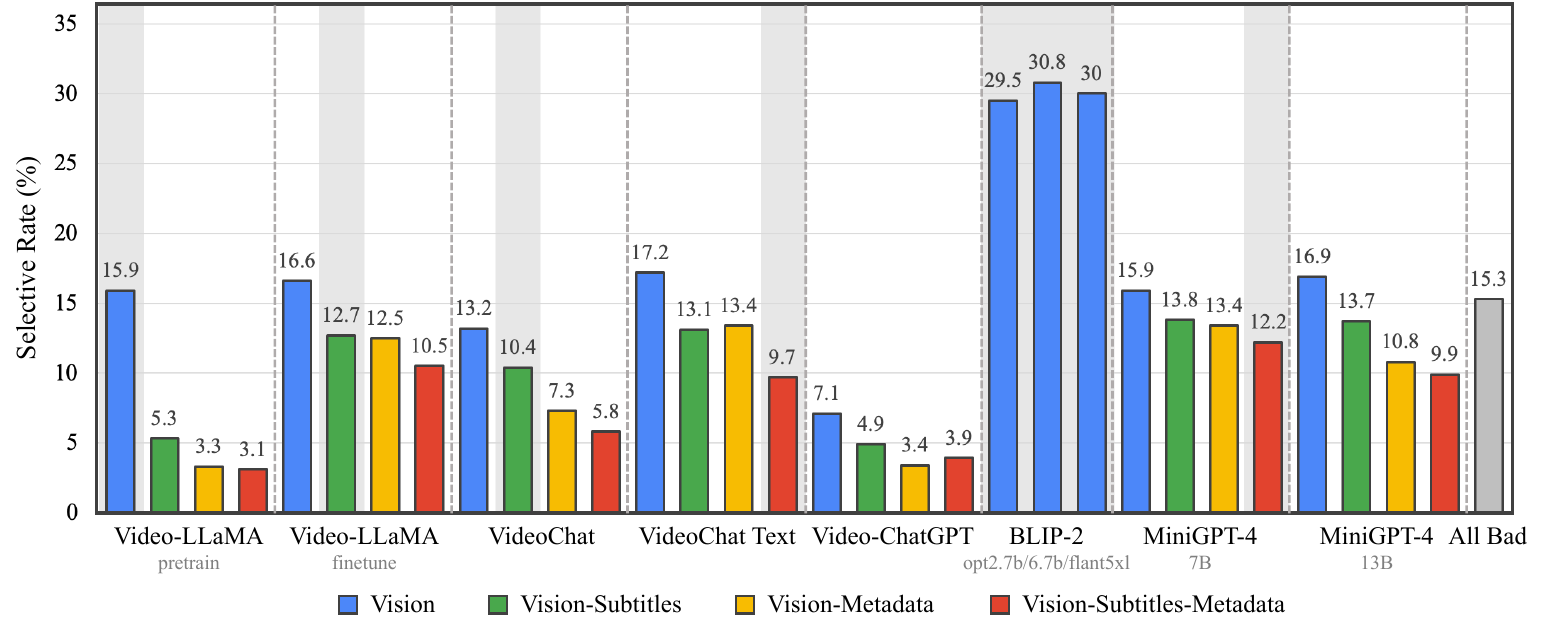}
    \vspace{-7mm}
    \mycaption{Ratio of an individual captioning model to predict a good caption}{
        % We show the percentage of an individual captioning model to predict a good caption.
        %
        Each bar represents an individual model and is colored by its input information.
        We highlight the 8 selected teacher models with {\color{gray} gray}.
        Note that we also report the ratio of ``All Bad'' at rightmost.
    }
    \vspace{-2mm}
    \label{fig:userstudy}
\end{figure}

\vspace{+1mm}
\noindent \textbf{Human evaluation.}
First, we conduct a user study by showing the output captions of each model to humans.
Specifically, we randomly sample 1K video clips and perform the inference of 31 captioning models on each video.
Next, the human annotators are asked to select ``\textit{every good caption}'', where a good caption is defined as: ``\textit{the caption cannot contain any wrong information and needs to cover the main action OR all of the main objects presented in the video.}''
If none of the captions is a good caption, the annotators are asked to select the ``All Bad'' option.
We randomly shuffle 31 captions to minimize the annotator's bias on the order of the captions.
Considering that a human is hard to focus on reading all 31 caption sentences at the same time, we split the captions into three groups.
The annotator will see the same video three times with at most 11 captions once.
We show the interface of this user study in Figure~\ref{fig:interface} and plot the results in Figure~\ref{fig:userstudy}.

\vspace{+1mm}
\noindent \textbf{Algorithm of model selection.}
In the second step, we collect a list of 8 captioning models as the representative to reduce the computation for large-scale captioning.
Intuitively, one may opt for the models exhibiting the top 8 performance.
Nonetheless, such behavior does not align with the philosophy of our captioning algorithm.
Specifically, our algorithm utilizes multiple cross-modality models to cover good captioning on various types of videos and only retrieves one best caption as the annotation for each video (as described in Section~\ref{sec:v2t_retrieval}).
Accordingly, we propose to use the set of models that can jointly cover a good caption for most video samples.
The algorithm starts by selecting the best-performing model (\ie, BLIP-2 with opt6.7b).
Next, we only consider the videos that the previously selected model(s) cannot generate a good caption and then greedily find the model that performs best on those videos.
We recursively collect the models under this mindset until we make the list of 8 captioning models.
The 8 selected models are highlighted in Figure~\ref{fig:userstudy}.

\vspace{+1mm}
\noindent \textbf{Additional findings.}
From Figure~\ref{fig:userstudy}, we can also observe that a single captioning model can predict a good caption for at most $\boldsymbol{30.8\%}$ of the videos.
In comparison, all 31 captioning can jointly predict at least one good caption for $\boldsymbol{84.7\%}$ of the videos (based on the ``All Bad'' ratio of $15.3\%$).
This fact supports our motivation to use multiple cross-modality teacher models to jointly predict the captions for a video.
Last but not least, according to our statistics, using 8 selected teacher captioning models can jointly predict a good caption for $\boldsymbol{76.8\%}$ of the videos which shows comparable performance with all 31 models while significantly reducing the computational requirements.

\section{Details of Fine-Grained Video-to-Text Retrieval: Dataset, Training, and Inference}
\label{app:retrieval}
In Section~\ref{sec:v2t_retrieval}, we mention that the available generic retrieval models~\cite{imagebind,unmaskedteacher} cannot pick the best caption from 8 candidates predicted by our teacher models.
The main reason is that all of the candidate captions are highly relevant to the video sample and require the model to discern subtle distinctions within each caption for optimal performance.
To better perform our ``fine-grained'' retrieval task, we first annotate a subset of video samples by manually selecting the best caption as detailed in Appendix~\ref{app:retrieval_dataset}.
Next, we finetune Unmasked Teacher~\cite{unmaskedteacher} (UMT) and run the inference of the model on all video samples respectively in Appendix~\ref{app:retrieval_finetune} and~\ref{app:retrieval_inference}.

\subsection{Collection of Dataset}
\label{app:retrieval_dataset}
We randomly sample 100K video samples from our dataset and ask human annotators to select ``\textit{the best caption}'' for each video.
At the beginning of the task, the annotator will read the task description as follows:

\textit{``You are presented with a short video clip and a set of textual summaries that describe this clip. Choose the textual summary that is the most faithful and descriptive of the content of the video clip. Imagine you are talking on the phone with your friend and you need to describe the video to him.''}

Note that this task is different from the user study in Appendix~\ref{app:user_study_all_good}, where a human is asked to select ``\textit{every good caption}''.
But, we also randomly shuffle the captions and provide an ``All Bad'' option if all of the captions contain wrong information.
We filter out $12,064$ videos with the ``All Bad'' option selected and split the dataset into $86,131$ and $1,805$ videos for training and validation.
We plot the selective rate of each teacher model on the validation set in Figure~\ref{fig:teacher} ({\color{NavyBlue}blue} bar).
%
% Notably, we compute the additional human agreement score on the validation set by having another group of annotators perform a second annotation round.
%
% The additional annotation shows $44.48\%$ R@1 accuracy compared to the original annotation.
%
% The results imply that the task is subjective when there is more than one good caption that performs equally well.

\subsection{Finetuning of Retrieval Model}
\label{app:retrieval_finetune}
We finetune Unmasked Teacher~\cite{unmaskedteacher} as the text retrieval model on the training set.
We use the larger model configuration, consisting of ViT-L/16~\cite{transformers} and BERTlarge~\cite{bert}, and initialize the model with the weights pretrained on 25M image-text and video-text pairs.
We follow the original codebase and only use the video-text contrastive (VTC) and video-text matching (VTM) loss functions for finetuning.
For VTC, we implement hard negative mining~\cite{albef,ifnd} which guides the model focusing on distinguishing the selected caption (\ie, the positive sample) and the other 7 captions (\ie, the hard negative samples).
Specifically, we set the training weights of the positive and hard negatives as $1$ while the weights of other negatives (\ie, captions from other videos) as $0.01$.
For the training videos, we randomly sample 12 video frames and apply $\mathrm{RandomResizedCrop}$ transformation with scale $[0.5, 1.0]$ to get the video with the resolution of $224 \times 224\mathrm{px}$.
We use the AdamW~\cite{adamw} optimizer with a learning rate of $2e^{-5}$, $\beta=[0.9, 0.999]$, and a weight decay of $0.02$.
We set the batch size of 32 and last the training for 10 epochs.
The model is funtuned on 8 Nvidia A100 GPUs (80GB).

\begin{figure}[t]
    \centering
    \includegraphics[width=\linewidth]{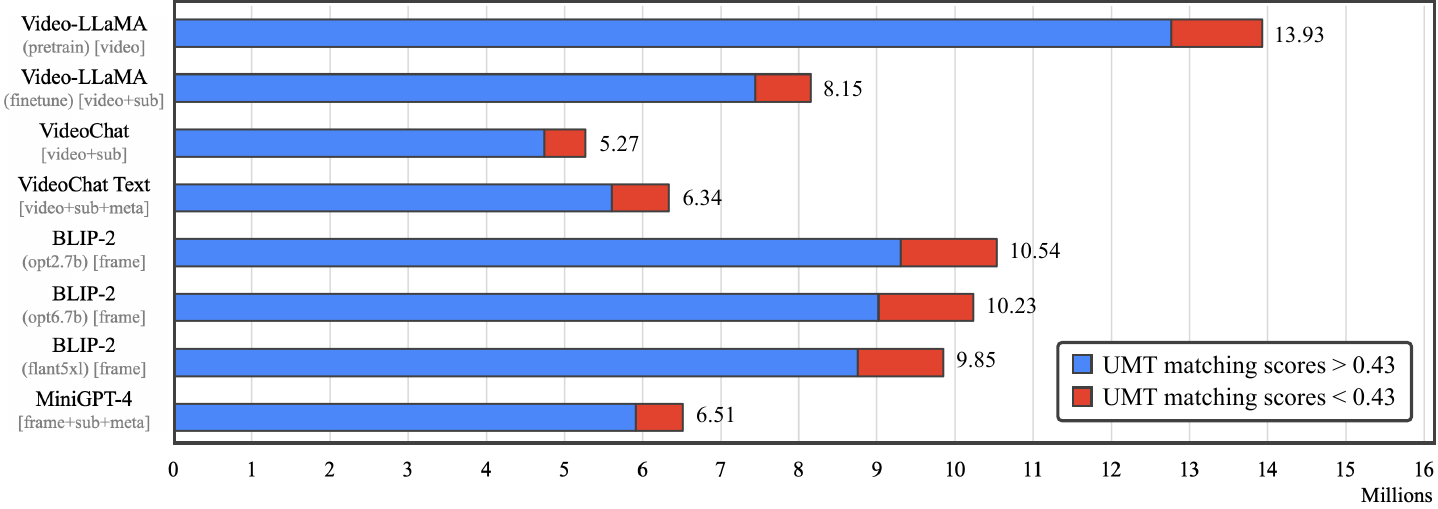}
    \vspace{-5mm}
    \mycaption{Distribution of the source teacher models of the captions in Panda-70M}{}
    \label{fig:teacher_70M}
\end{figure}

\subsection{Inference of Retrieval Model on Panda-70M}
\label{app:retrieval_inference}
\begin{figure}[t]
    \centering
    \includegraphics[width=.9\linewidth]{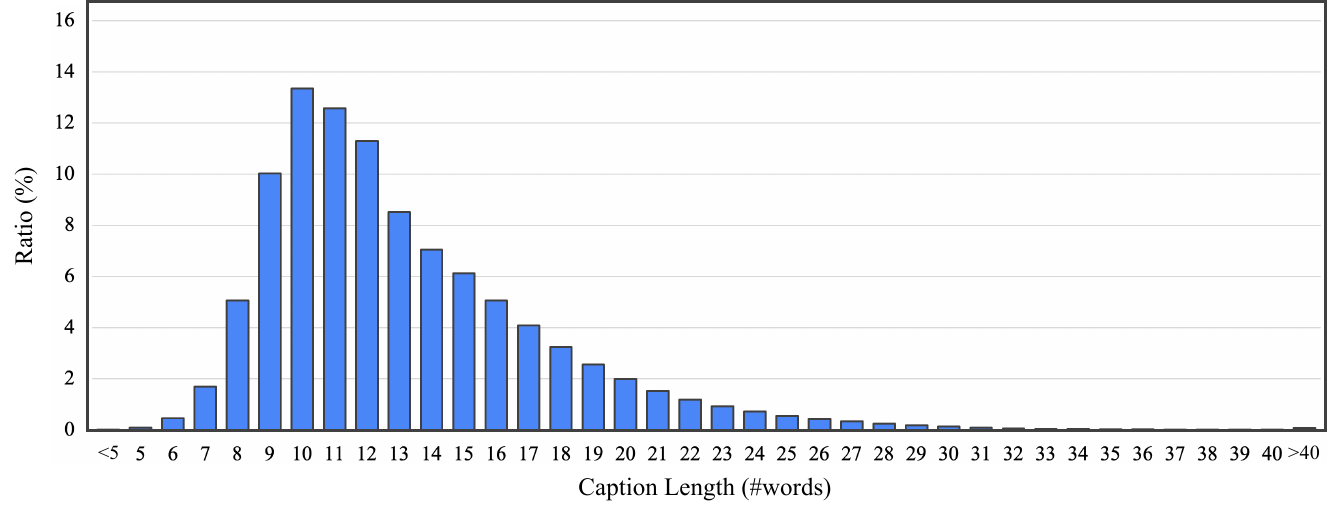}
    \vspace{-3mm}
    \mycaption{Distribution of caption length of Panda-70M}{}
    %\vspace{-3mm}
    \label{fig:caption_length}
\end{figure}

\begin{figure}[t]
    \centering
    \includegraphics[width=.9\linewidth]{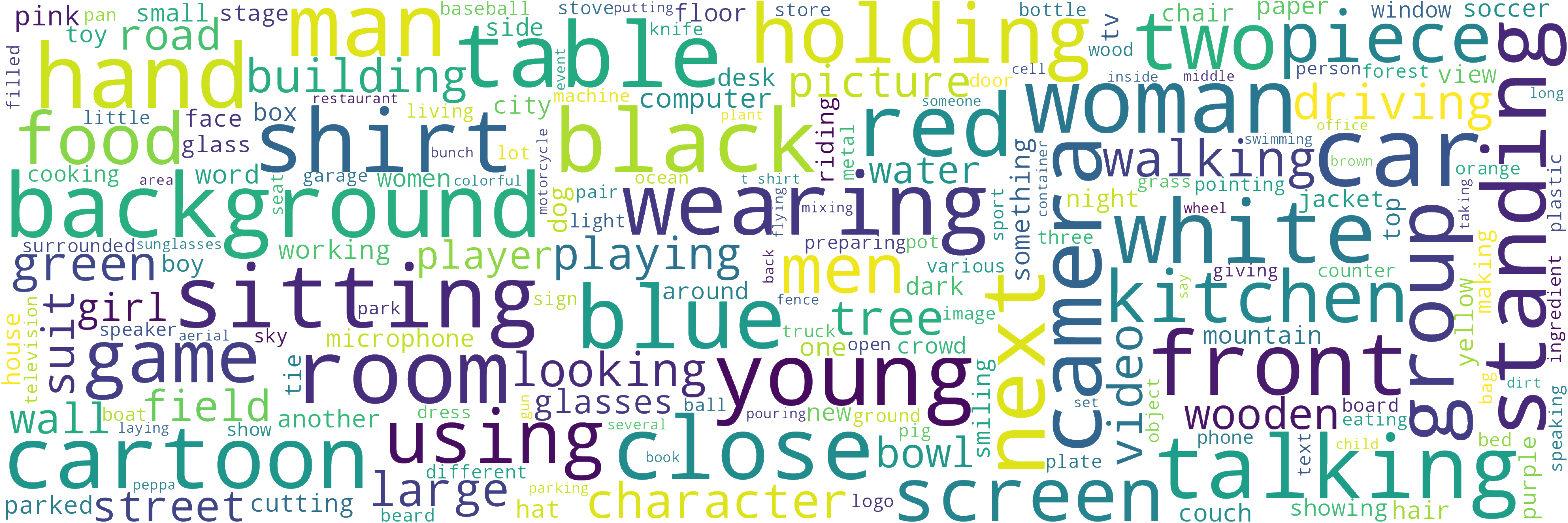}
    %\vspace{-7mm}
    \mycaption{Word cloud of 100K caption samples in Panda-70M}{}
    \vspace{-1mm}
    \label{fig:worldcloud}
\end{figure}

With the finetuned UMT, we automatically retrieve the best caption as the annotation for all 70M videos.
We illustrate the distribution of the finetuned UMT's selection in Figure~\ref{fig:teacher_70M} and the caption length in Figure~\ref{fig:caption_length}.
We also plot the word cloud of the randomly sampled 100K caption annotations in Figure~\ref{fig:worldcloud} to highlight the rich content within the annotated captions.

In addition to the retrieval result, UMT also predicts a matching score for the video-text pair.
In practice, we find the score is highly correlated to the alignment of the contents within the video-text pair.
A score higher than $0.43$ usually represents a strong association between the video and the caption.
Numerically, $89.6\%$ of the samples in Panda-70M have matching scores higher than $0.43$.

\section{Details of Student Captioning Model: Architecture and Training}
\label{app:student}

\subsection{Model Architecture}
Figure~\ref{fig:architecture} shows the architecture of the student captioning model.
The model includes a vision branch and a text branch for additional subtitle and metadata inputs.

The vision branch shares the same design as Video-LLaMA~\cite{videollama}.
Specifically, given an 8-frame video with the resolution of $224 \times 224\mathrm{px}$, a visual encoder first individually encodes each video frame into multiple frame-level features with the dimension of $32 \times 768$.
The visual encoder is composed of a frozen pretrained visual encoder, including a ViT-G/14 from EVA-CLIP~\cite{eva} and a Q-former~\cite{blipv2}.
Subsequently, the temporal fusion module aggregates multiple frame-level features into a single $32 \times 768$ video representation.
The module includes a position embedding layer to inject temporal information into video frames and a video Q-former to fuse the frame-level features.
Finally, the model projects the video representation into a $32 \times 4096$ feature by a linear layer.

For the text branch, given a prompt with an arbitrary length, the model first tokenizes the prompt and embeds each token into a feature vector with $4096$ length by a pretrained embedding layer~\cite{vicuna}.
Considering that the number of token embedding might be large for a longer prompt and the information of the prompt might not well align with the video content, we then design a text Q-Former to extract a fixed and shorter length of text embedding and at the same time, better bridge the feature of the input video and text prompt.
Specifically, the text Q-Former takes the inputs of the $32 \times 4096$ video representation as the queries and multiple token embedding as the key and value.
The module then outputs a $32 \times 4096$ text representation.
Finally, we combine the multimodal inputs by concatenating the text and video representations in sequence to get a $64 \times 4096$ feature and input it to the LLM to predict the video caption.

\subsection{Training Details}
The training data includes a video-caption pair and additional text information (\ie, the metadata and subtitles).
For the video data, we randomly sample 8 frames and apply the same video reading algorithm as in Appendix~\ref{app:retrieval_finetune}.
For the text branch, we embed the extra text information into the prompt. 
To learn a captioning model that can take both video-only and video-text inputs, we drop part of the text inputs at random.
Formally, we use the prompt template in Figure~\ref{fig:prompt} and employ the metadata or/and subtitles information with the probability of $0.5$ (the sampling for metadata and subtitles are independent).

We use the AdamW~\cite{adamw} optimizer.
The learning rate is initialized as $1e^{-6}$ and linearly warmed up to $1e^{-4}$ within the first 2,500 steps and gradually decreased to $5e^{-5}$ based on cosine annealing strategy~\cite{sgdr}.
We set $\beta=[0.9, 0.99]$ and use a weight decay of $0.05$.
We train the model on the whole Panda-70M with a batch size of 48 and last the training for 300K steps.
The model is trained on 48 Nvidia A100 GPUs (80GB).
\section{Visualization of Panda-70M Dataset}
\label{app:visualization}
In the following subsections, we visualize video-text pairs in Panda-70M by category.
%
% More video samples can be found on the Supplementary website.

\subsection{Category: Animal}
\begin{figure}[h]
    \centering
    \vspace{-2mm}
    \includegraphics[width=\linewidth]{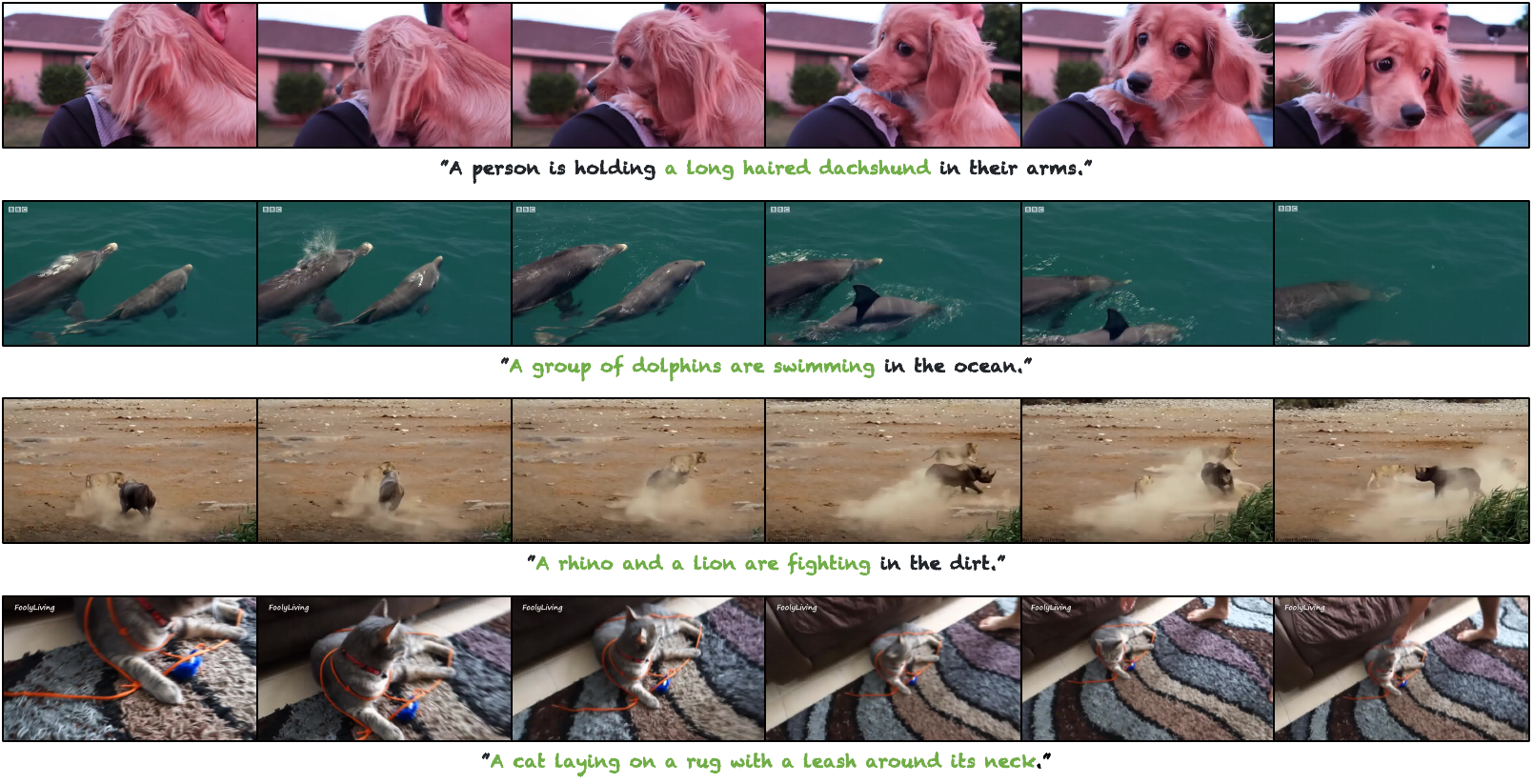}
    \vspace{-7mm}
\end{figure}

\clearpage
\subsection{Category: Scenery}
\begin{figure}[h]
    \centering
    \vspace{-2mm}
    \includegraphics[width=\linewidth]{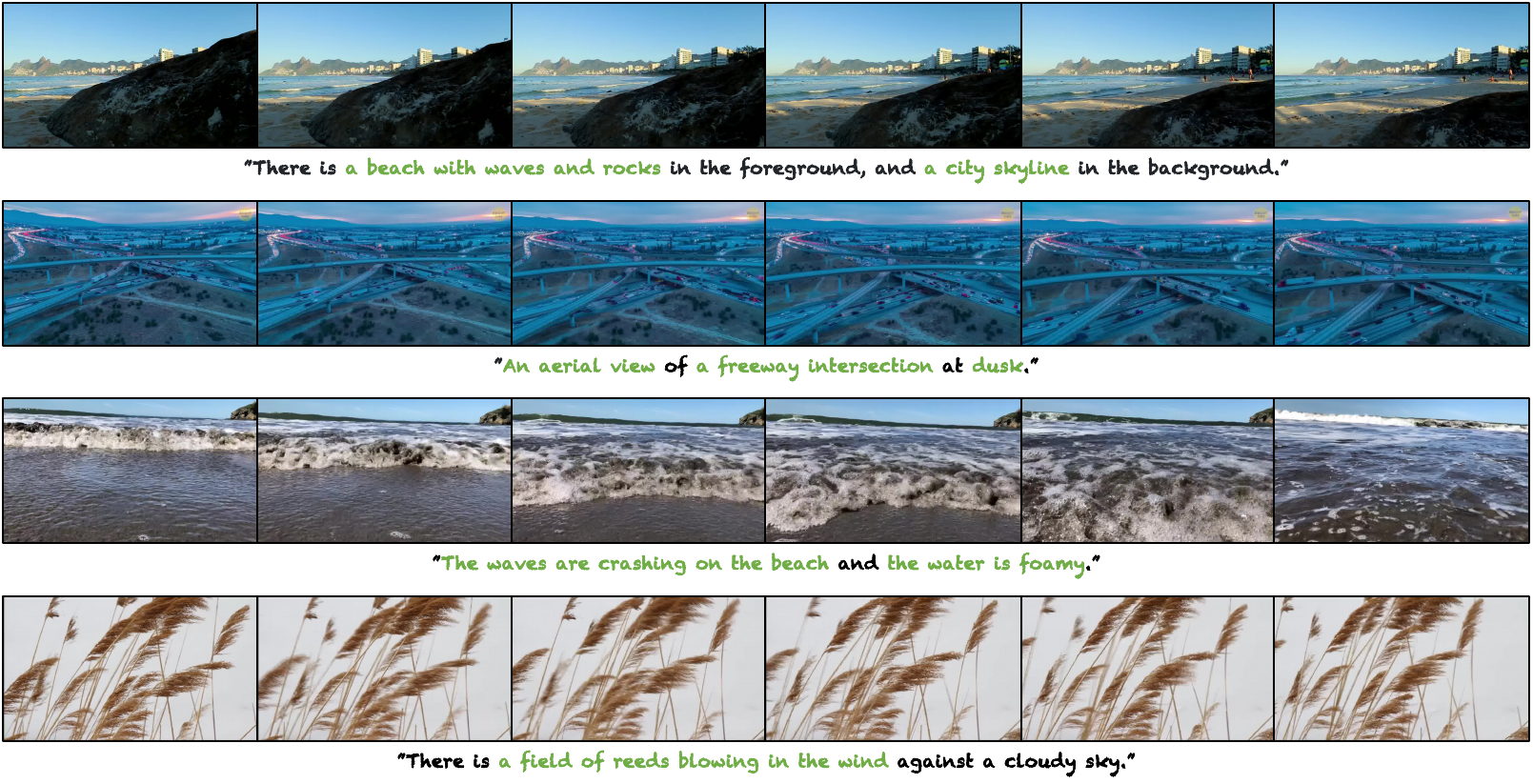}
    \vspace{-7mm}
\end{figure}
\subsection{Category: Food}
\begin{figure}[h]
    \centering
    \vspace{-2mm}
    \includegraphics[width=\linewidth]{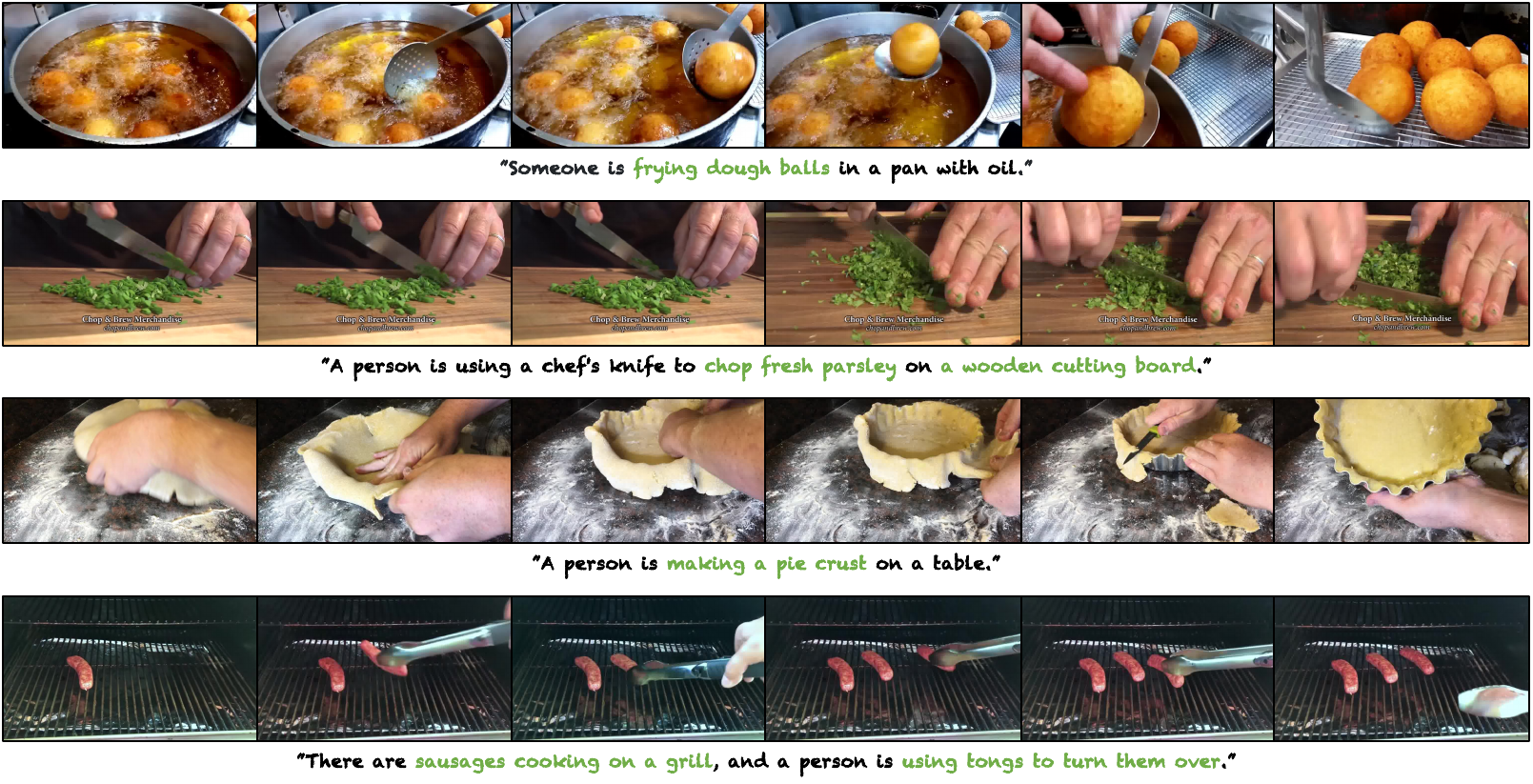}
    \vspace{-7mm}
\end{figure}

\clearpage
\subsection{Category: Sports Activity}
\begin{figure}[h]
    \centering
    \vspace{-2mm}
    \includegraphics[width=\linewidth]{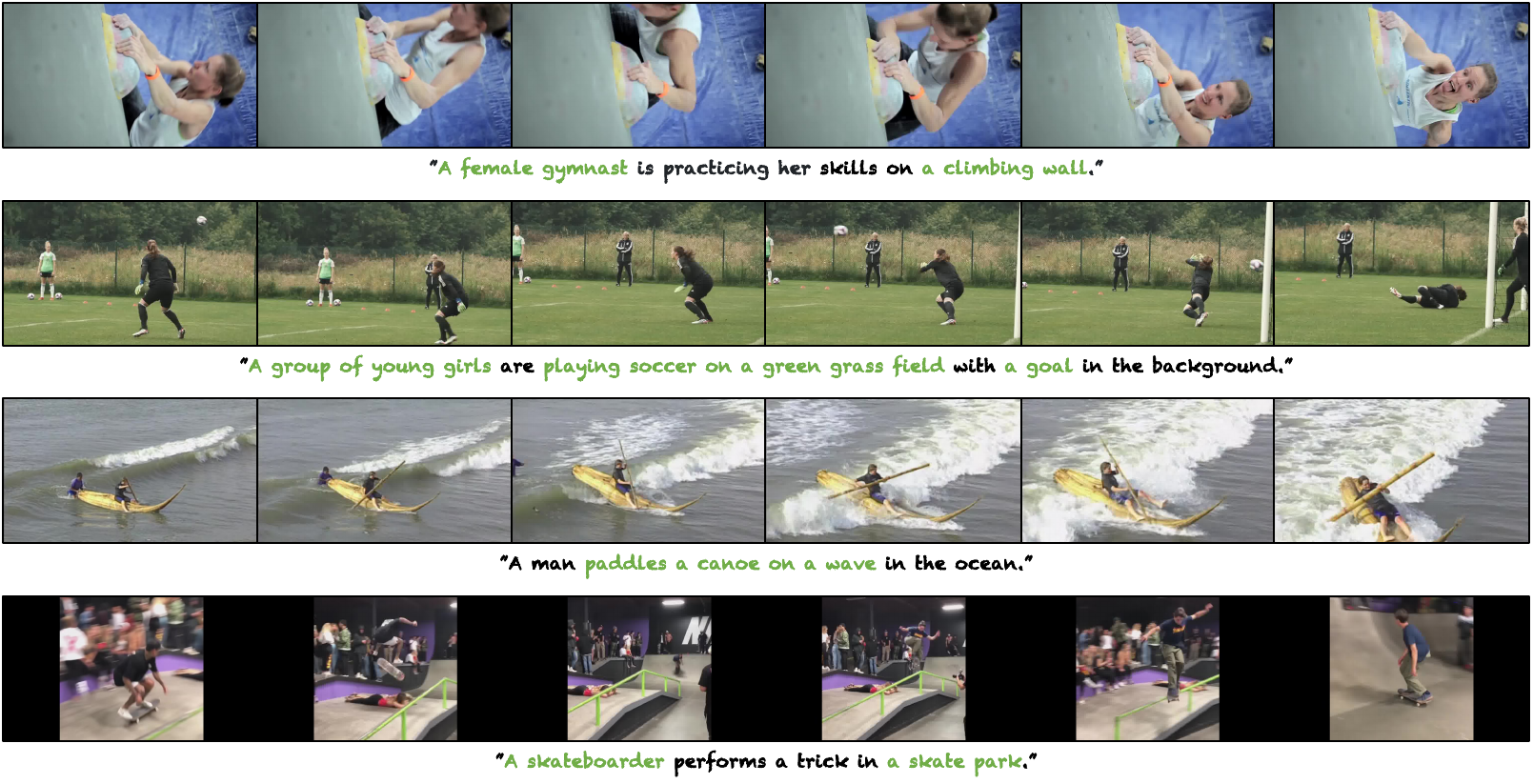}
    \vspace{-7mm}
\end{figure}
\subsection{Category: Vehicles}
\begin{figure}[h]
    \centering
    \vspace{-2mm}
    \includegraphics[width=\linewidth]{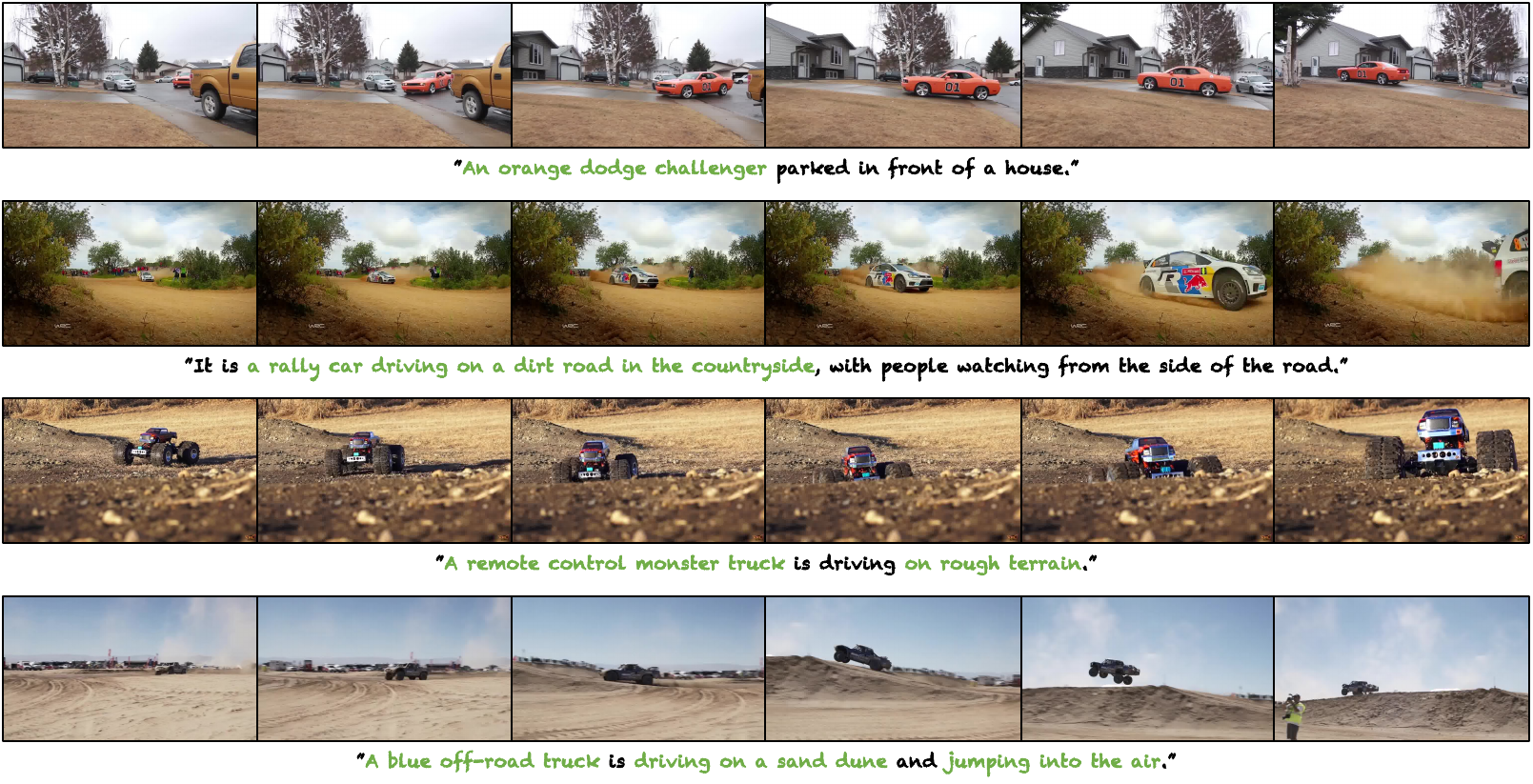}
    \vspace{-7mm}
\end{figure}

\clearpage
\subsection{Category: Tutorial and Narrative}
\begin{figure}[h]
    \centering
    \vspace{-2mm}
    \includegraphics[width=\linewidth]{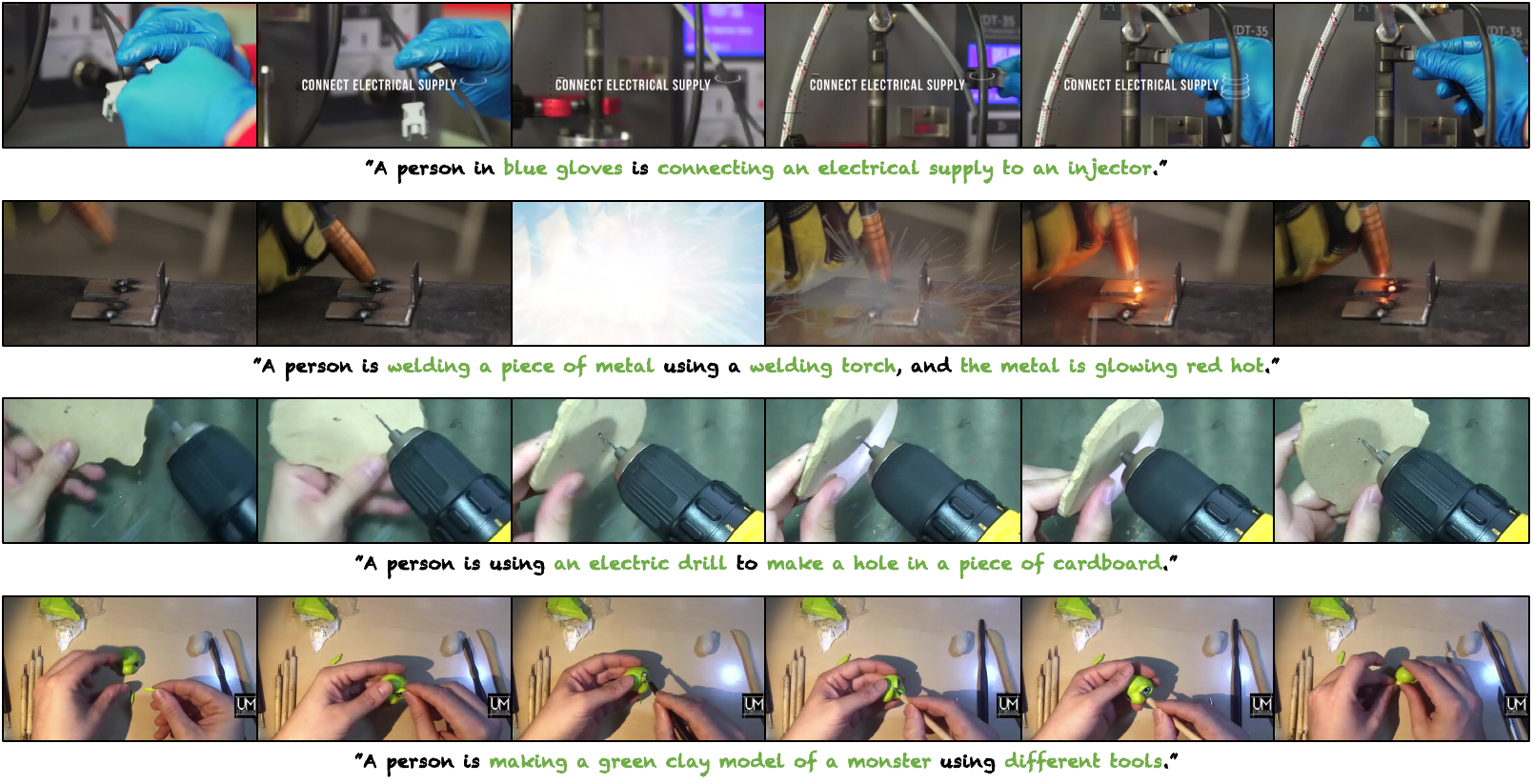}
    \vspace{-7mm}
\end{figure}
\subsection{Category: News and TV Shows}
\begin{figure}[h]
    \centering
    \vspace{-2mm}
    \includegraphics[width=\linewidth]{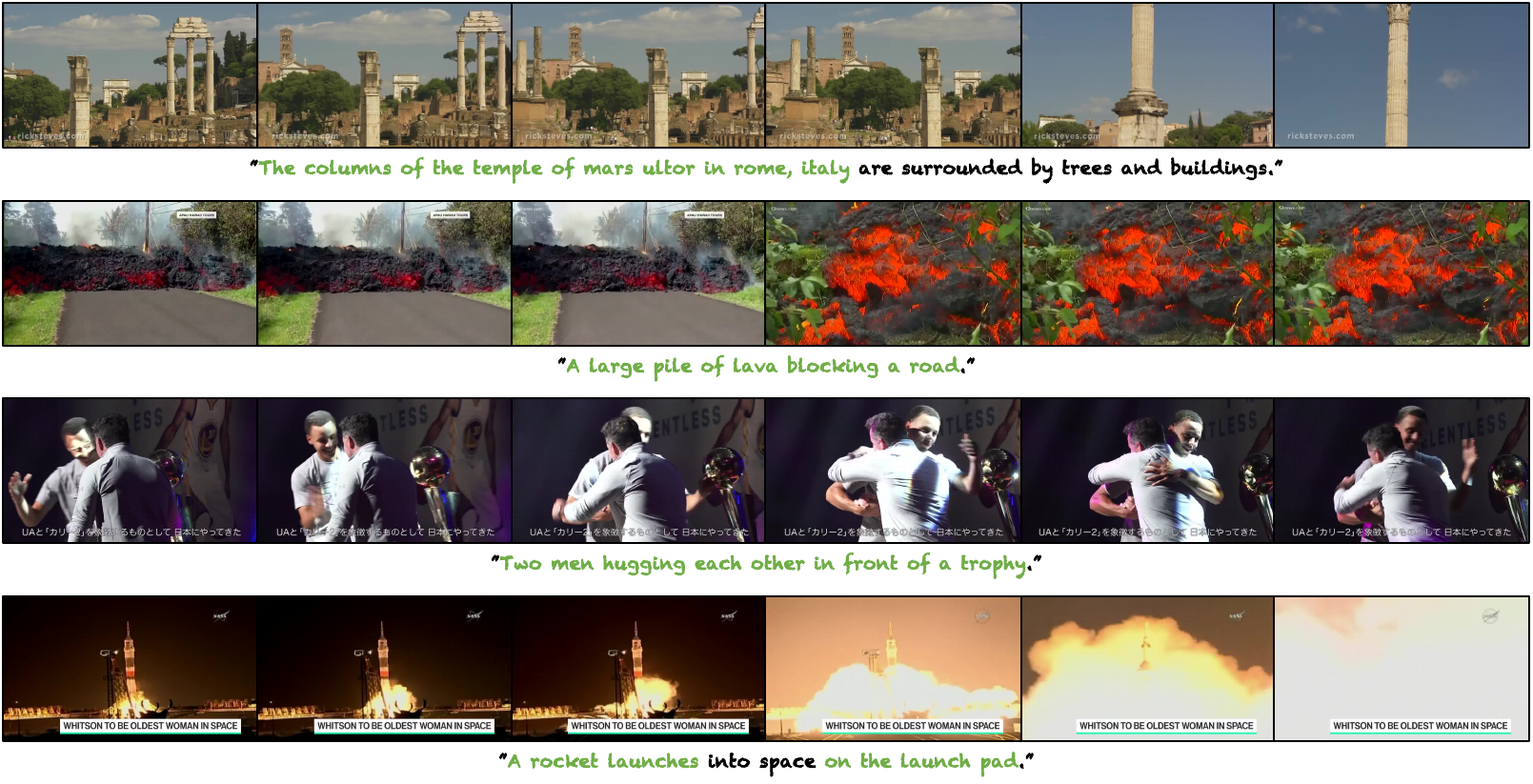}
    \vspace{-7mm}
\end{figure}

\clearpage
\subsection{Category: Gaming and 3D Rendering}
\begin{figure}[h]
    \centering
    \vspace{-2mm}
    \includegraphics[width=\linewidth]{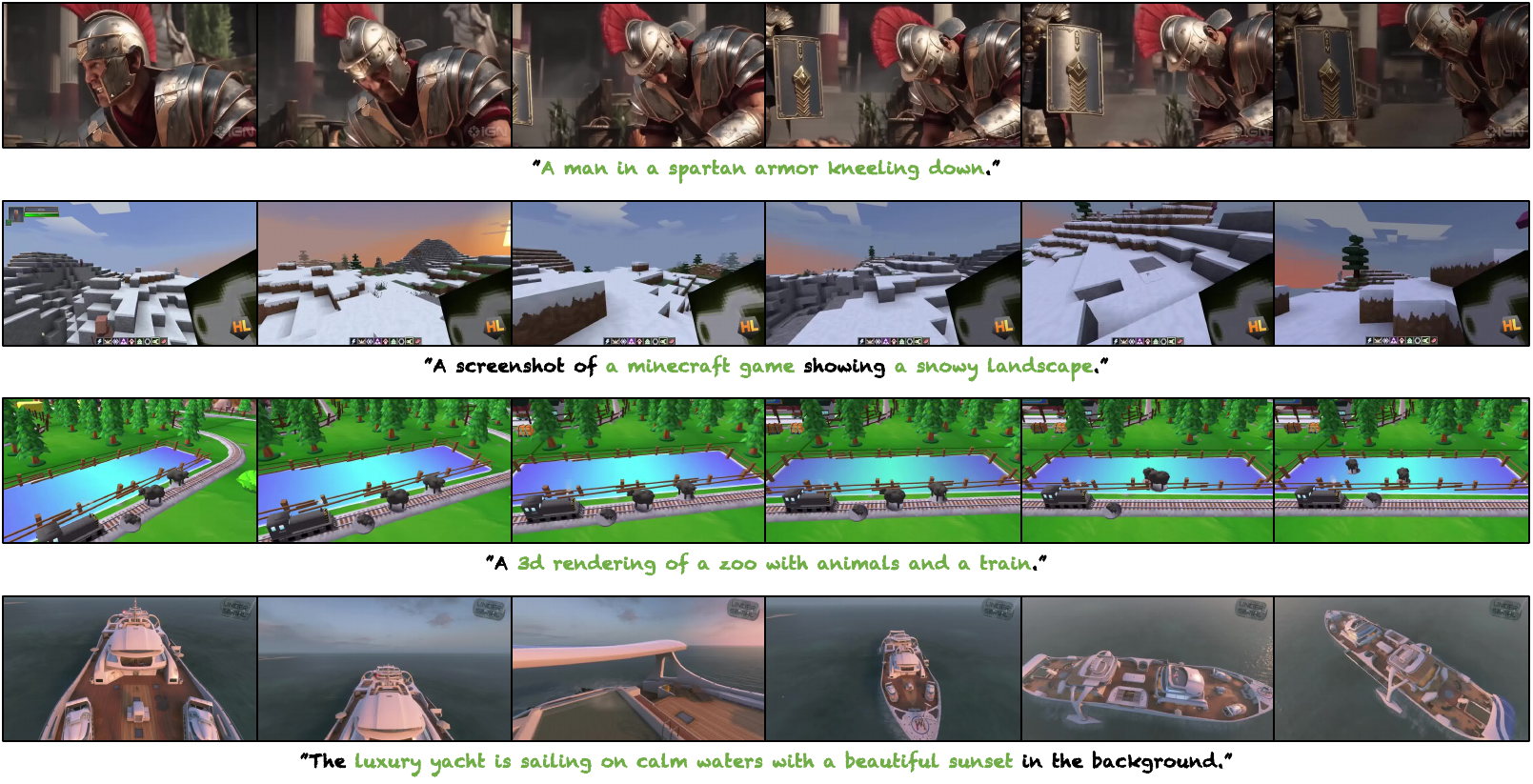}
    \vspace{-7mm}
\end{figure}

\end{document}